\pdfoutput=1

\documentclass[11pt]{article}

\usepackage[final]{acl}

\usepackage{times}
\usepackage{latexsym}

\usepackage[T1]{fontenc}

\usepackage[utf8]{inputenc}

\usepackage{microtype}

\usepackage{inconsolata}

\usepackage{graphicx}
\usepackage{amsmath}
\usepackage{amssymb}
\usepackage{booktabs}
\usepackage{subcaption}
\usepackage{float}
\usepackage{multirow}
\usepackage{array}
\usepackage{makecell}
\usepackage{graphicx}
\usepackage{listings}
\usepackage{mathtools}
\usepackage{enumitem}
\usepackage{hyperref}

\usepackage{cleveref}
\newcommand{\Scref}[1]{\S\ref{#1}}

\usepackage{xspace}
\newcommand{\method}{\textsc{V-SEAM}\xspace} 

\usepackage{xcolor}
\newif\ifdraft\drafttrue

\usepackage{xspace}
\newcommand{\framework}{V-SEAM\xspace}

\usepackage{amsmath,amsfonts,bm}

\def\eqref#1{equation~\ref{#1}}

\def\1{\bm{1}}

\def\mA{{\bm{A}}}

\def\mH{{\bm{H}}}

\DeclareMathAlphabet{\mathsfit}{\encodingdefault}{\sfdefault}{m}{sl}
\SetMathAlphabet{\mathsfit}{bold}{\encodingdefault}{\sfdefault}{bx}{n}

\def\gB{{\mathcal{B}}}

\def\gD{{\mathcal{D}}}

\def\gH{{\mathcal{H}}}
\def\gI{{\mathcal{I}}}

\def\sR{{\mathbb{R}}}

\title{V-SEAM: Visual Semantic Editing and Attention Modulating for Causal Interpretability of Vision-Language Models}

\author{
    Qidong Wang$^1$, Junjie Hu$^2$, Ming Jiang$^2$\thanks{corresponding author} \\
    $^1$Tongji University, $^2$University of Wisconsin-Madison\\
    \texttt{wang\_qidong@tongji.edu.cn}, \texttt{\{junjie.hu, ming.jiang\}@wisc.edu}
}

\begin{document}
\maketitle

\begin{abstract}

Recent advances in causal interpretability have extended from language models to vision-language models (VLMs), seeking to reveal their internal mechanisms through input interventions. While textual interventions often target semantics, visual interventions typically rely on coarse pixel-level perturbations, limiting semantic insights on multimodal integration. In this study, we introduce \framework, a novel framework that combines \textbf{V}isual \textbf{S}emantic \textbf{E}diting and \textbf{A}ttention \textbf{M}odulating for causal interpretation of VLMs. V-SEAM enables concept-level visual manipulations and identifies attention heads with positive or negative contributions to predictions across three semantic levels: objects, attributes, and relationships. We observe that positive heads are often shared within the same semantic level but vary across levels, while negative heads tend to generalize broadly. Finally, we introduce an automatic method to modulate key head embeddings, demonstrating enhanced performance for both LLAVA and InstructBLIP across three diverse VQA benchmarks. Our data and code are released at: \url{https://github.com/petergit1/V-SEAM}.

\end{abstract}
\section{Introduction}
\label{sec:introduction}

Vision-language models (VLMs)  have become a vital infrastructure for multimodal understanding and generation, powering a variety of downstream applications, such as visual question answering (VQA) ~\cite{liu2023visual,dai2023instructblip,liu2024improved}, image captioning ~\cite{zhang2021vinvl,wang2024cogvlm}, and navigation ~\cite{anderson2018vision,zhou2024navgpt}. Despite these models' impressive performance, VLMs' internal mechanisms remain underexplored, risking their trustworthiness in real deployments. To address this gap, researchers have increasingly focused on interpreting VLMs, with particular emphasis on causal intervention methods given their dual benefits: causally unraveling model behaviors and providing systematic pathways for model improvement, like model editing ~\cite{zhao2024explainability,lin2025survey}.

\begin{figure}[t]
    \centering
    \includegraphics[scale=0.28]{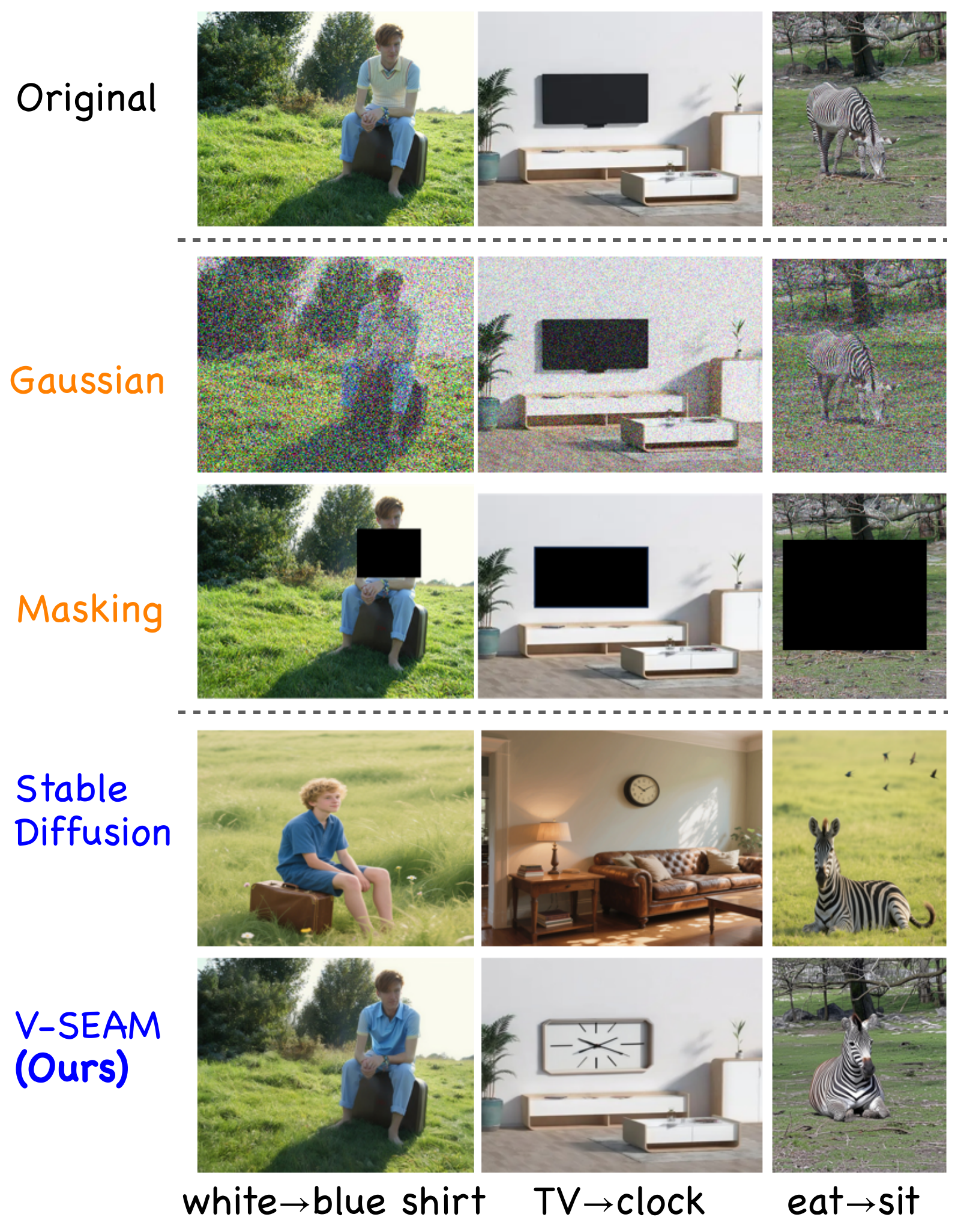}
    \caption{\small
    An example of visual intervention comparisons: Visual \textcolor{orange}{non-semantic} vs. \textcolor{blue}{semantic} interventions.
    }
    \vspace{-16pt}
    \label{fig:intro_visual_comparison}
\end{figure}

Existing work on causal interpretability primarily focuses on large language models (LLMs)~\cite{vig2020investigating,meng2022locating,geva2023dissecting,zhao2024explainability}. Recently, some studies have adapted this strategy to VLMs by perturbing both visual and text inputs ~\cite{palit2023towards,basuunderstanding,golovanevsky2025vlms}. However, unlike textual interventions (e.g., token replacement), which focus on fine-grained semantic changes ~\cite{basuunderstanding,zhang2023towards}, visual interventions, such as image-wide Gaussian noise ~\cite{palit2023towards} and random visual masking ~\cite{neo2024towards}, typically emphasize holistic disruptions at the pixel level (see Figure~\ref{fig:intro_visual_comparison}). \citet{zhang2023towards} shows that this intervention gap leads to inconsistent interpretations of model mechanisms and can even produce deceptive outcomes, causing uncertainty with VLM analysis. Although the latest works propose semantic-based visual intervention by either retrieving similar image pairs or generating adversarial images via stable diffusion under the guidance of text instructions ~\cite{golovanevsky2025vlms}, this approach, as shown in Figure~\ref{fig:intro_visual_comparison}, may introduce unintended alterations to the original image, such as changes to the background or the main object. Consequently, it risks incorporating extraneous disturbances into model analysis.

In this study, we introduce \framework, a fine-grained, semantics-based interpretability framework for VLMs. Motivated by the intuition that ideal visual interventions should alter only the image components relevant to a specific semantic property while preserving the rest, we propose \textit{\textbf{V}isual \textbf{S}emantic \textbf{E}diting} to manipulate visual inputs at three semantic categories: (1) objects, (2) object attributes, and (3) object-object relationships. Building on this, we develop \textit{\textbf{A}ttention \textbf{M}odulating}, a causal method that identifies key attention heads driving predictions at each semantic level and modulates their embeddings for model improvement.

Using \framework on the visual question answering (VQA) task, we empirically investigate two popular VLMs, i.e., LLAVA and InstructBLIP, and identify four key findings: (1) VLMs generally emphasize object-level cross-modal alignment in early layers (e.g., layers 5–10), shifting to attributes and relationships in later layers (e.g., layers 9–15); (2) attention modules primarily capture salient semantic cues (e.g., color) that inform predictions, while MLP blocks play a key role in producing the final decisions; (3) for individual attention heads, we identify both positive heads that facilitate correct predictions and negative heads that introduce misleading signals. Positive heads are generally shareable within the same semantic categories but differ across categories, whereas negative heads tend to generalize across semantics; and (4) by rescaling key attention head embeddings, we achieve improved VQA accuracy for both LLAVA and InstructBLIP on three diverse benchmarks.

Overall, our main contributions are as follows:

\begin{itemize}[leftmargin=13pt]
    \item We introduce \framework, a novel mechanistic interpretability framework that enables concept-level visual semantic editing and attention modulating to causally trace key components in VLMs for object-, attribute-, and relation-level visual understanding.

    \item We uncover a variety of novel insights into how VLMs process fine-grained semantic information from visual and textual inputs.

    \item We apply \framework to improve model performance on VQA tasks by amplifying positive attention heads and suppressing negative ones through embedding rescaling. Experiments on three diverse VQA tasks demonstrate consistent improvements in VLM performance.

\end{itemize}

\section{Related Work}  
\label{sec:related}

\paragraph{Vision-Language Models}
Existing VLMs can be broadly categorized into two groups based on their strategies for integrating visual and textual information. One group employs cross-attention mechanisms to embed visual features into textual ones~\cite{alayrac2022flamingo,li2023blip2}, while the other maps visual and textual features through a projection layer~\cite{liu2023visual, bai2023qwenvl,zhu2024minigpt4,chen2024internvl,li2024llava}. Since projection-based VLMs typically achieve stronger performance, prior interpretability work has predominantly focused on this group~\cite{bai2023qwenvl,liu2023visual}.  With the goal of offering a more comprehensive causal interpretation, we examine both groups of VLMs in this study.

\paragraph{Multimodal Interpretability}

Existing research on the interpretability of vision-language models (VLMs) can be broadly categorized into two complementary approaches: data-driven methods~\cite{sood2023multimodal,xing2025large} and model structure–oriented analyses. Data-driven methods typically employ saliency-based techniques to associate specific input regions, e.g., pixels or visual tokens, with model outputs, thereby offering insights into which inputs most strongly influence predictions. Unlikely, structure-based analyses aim to interpret VLMs by examining their internal components and mechanisms, ranging from the study of hierarchical representations across layers~\cite{palit2023towards,zhang2024redundancy}, the functional roles of distinct modules~\cite{ben2024lvlm,basuunderstanding,chen2025spatial}, to the contribution and specialization of attention heads~\cite{golovanevsky2025vlms}, and the behavior of individual neurons or neuron groups~\cite{gandelsman2023interpreting}. Our work falls into this group, emphasizing both hierarchical representations and attention mechanisms within VLMs.

Within structure-based analyses, there are two major strands. One is correlation-based interpretations, diagnosing what kinds of semantic information are encoded across VLM layers and components. A common technique is probing, where classifiers are trained on hidden states to decode semantic properties~\cite{alain2016understanding,hendricks2021probing}. One major issue with this method is its sensitivity to probe design and vulnerability to spurious correlations that obscure true model representations~\cite{belinkov2022probing, bilodeau2024impossibility}. To address this, alternative methods have been developed. For example, TextSpan~\cite{gandelsman2023interpreting} and Logit Lens~\cite{neo2024towards} aim to align neurons or visual tokens with discrete, human-interpretable concepts, offering a more direct view into model internals. Neuron attribution methods such as MINER~\cite{huang2024miner} and MMNeuron~\cite{huo2024mmneuron} identify neurons selectively activated by specific tasks or semantics. While these provide fine-grained insights into VLM organization, their interpretability is limited by neuronal polysemanticity, where single neurons may encode multiple, entangled concepts~\cite{liu2025mechanistic}. In parallel, sparse autoencoder techniques show promise in disentangling representations in LLMs and ViTs~\cite{cunningham2023sparse}, but face challenges in VLMs due to modality inconsistency and instability.

The second strand emphasizes causal tracing, investigating the causal effect of input interventions on model components~\cite{vig2020investigating,geva2023dissecting}. For example, \citet{basuunderstanding} found that early VLM layers encode factual knowledge, while \citet{palit2023towards} highlighted the role of deeper layers. \citet{golovanevsky2025vlms} further analyzed modality-specific attention heads. However, existing interventions often rely on coarse perturbations (e.g., full-image replacement or noise), which can introduce artifacts and obscure semantic causality \cite{zhang2023towards}. To address this, we propose a vision-centric causal interpretability framework that supports fine-grained interventions. Our method combines semantic manipulations on image regions with activation patching on image and text tokens, enabling analysis of how fine-level semantic types (attributes, objects, relations) are processed and aligned across layers and modalities. We further perform attention-head-level interventions to identify “positive” and “negative” heads sensitive to distinct semantics, extending prior modality-specific attention analysis \cite{golovanevsky2025vlms}.
\section{Preliminary}
\label{sec:preliminary}

We first define the notations of data and models, then describe the \textit{activation patching}~\cite{meng2022locating} method (a.k.a. \textit{causal tracing}) in the context of VLMs~\cite{golovanevsky2025vlms}.

\paragraph{Data and Models.} We consider a VQA dataset $\gD$ consisting of $N$ question-image-answer triples. Each triple contains a question $x$ about an image $z$, and a single-token answer $y$. Each image $z$ is annotated with a list of objects $\gB$, including their text labels and bounding box coordinates, and the object-object relation edges. We define a transformer-based VLM as $f_\theta$, where its parameters $\theta=\{\theta_\text{in}, \theta_\text{out}\} \cup \{ \theta_\text{att}^{l}, \theta_\text{mlp}^l\}_{l=1}^L$ include the vision and text input encoding layer $\theta_\text{in}$, the output projection layer $\theta_\text{out}$ and $L$ transformer layers, each composed of self-attention $\theta_\text{att}^l$ and MLP $\theta_\text{mlp}^l$ modules. We denote the output logit of the correct answer $y$ to a given input $(x,z)$ as $\ell(x, z, y)\in\sR$.

\begin{figure*}[ht]
    \centering
    \includegraphics[width=\linewidth]{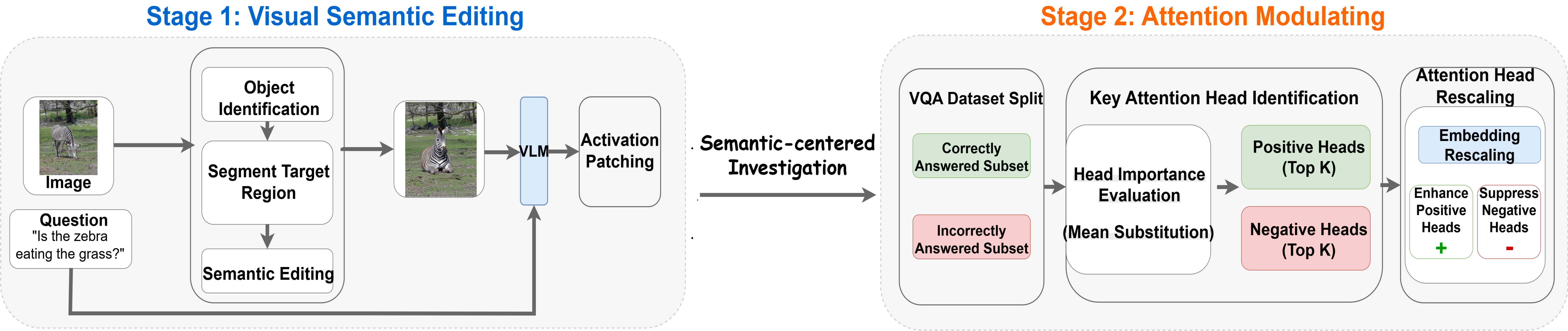}
    \caption{Our proposed semantic-level causal interpretability framework. The pipeline starts from semantic-guided perturbation of visual regions (attributes, objects, or relations), followed by vision-language model reasoning, and ends with causal analysis through activation patching and attention head ablation.}
    \label{fig:framework}
    \vspace{-1mm}
\end{figure*}

\paragraph{Activation Patching in VLMs.} Given a VQA triple $(x, z, y)\in\gD$, the method first modifies the original clean image $z$ to obtain a corrupted image $\tilde{z}$, denoted as $\tilde{z}$. It then feeds $(x, z)$ and $(x, \tilde{z})$ independently into the VLM, obtaining layerwise embeddings after self-attention and MLP modules~denoted as $\mH_{\tau}^{l}$, $\tilde{\mH}_{\tau}^{l}\in\sR^{T\times d}$ respectively, where $\tau\in\{\text{att},\text{mlp}\}$, $T$ is the input sequence length, $d$ is the hidden dimension. The model's output logits for the correct answer $y$ are thus denoted as $\ell(x, z, y), \ell(x, \tilde{z}, y) \in \sR$ respectively. We further identify the set of indices of the corrupted tokens in the combined input sequence $(x, \tilde{z})$ as $\gI$.

An activation patching operation is then applied to modify $\tilde{\mH}_\tau^l$, where $\tau\in\{\text{att},\text{mlp}\}$. Specifically, the embeddings of the corrupted tokens in $\tilde{\mH}_\tau^l$ are selectively replaced with their clean counterparts in $\mH_\tau^l$, based on the indices $\gI$ of the corrupted tokens.
\begin{align}
    \hat{\mH}_{\tau}^l \leftarrow \text{Patch}(\mH^l_{\tau}, \tilde{\mH}^l_{\tau}, \gI) 
\end{align}

The patched embedding $\hat{\mH}^l_\tau$ is fed through the remaining VLM layers to obtain the patched logit for the correct answer $y$, denoted $\hat{\ell}^{l}_\tau(x, \tilde{z}, y)\in \sR$. We then compute its difference from the corrupted logit to measure the effect of patching:
\begin{align}
    \Delta \ell^l_\tau(x, \tilde{z}, y) = \hat{\ell}^{l}_\tau(x, \tilde{z}, y) - \ell(x, \tilde{z}, y).
\end{align}
Intuitively, a large positive value of $\Delta \ell^l_\tau(x, \tilde{z}, y)$ indicates that patching the corrupted embeddings after module $\tau$ at layer $l$ has a beneficial effect on predicting the correct answer for a given VQA triple. This allows us to identify modules that are critical for the model's performance. To improve estimation stability, we compute the final causal score for a VLM module as the averaged logit difference across all the $N$ triples in $\gD$:
\begin{align}
    s(\tau, l) = \frac{1}{N} \sum_{(x, z, y)\in\gD} \Delta \ell_\tau^l(x, \tilde{z}, y).
\end{align}

\section{\method}
\label{sec:interpretation_method}

Figure~\ref{fig:framework} provides an overview of our proposed two-stage framework. In the first stage, \method applies visual semantic editing on local image regions and layerwise activation patching to identify key modules and their functional roles in VLMs for visual semantic understanding (\Scref{sec:seam:semantic_editing}). Building on these insights, the second stage focuses on attention modulation, identifying critical attention heads that positively or negatively affect visual semantic understanding and prediction, and enabling targeted intervention via attention head rescaling to enhance VLM performance (\Scref{sec:seam:attention_masking}).

\subsection{Visual Semantic Editing}
\label{sec:seam:semantic_editing}

As shown in Figure~\ref{fig:intro_visual_comparison}, current visual intervention methods often rely on coarse perturbations that lack visual semantic precision, making it infeasible to isolate the causal contributions of specific visual semantic elements such as color, object identity, or spatial relations. Moreover, strong additive perturbations can push input data out of the model's training data distribution, leading to unreliable or misleading attributions~\cite{zhang2023towards}. To address these limitations, we propose a \textit{visual semantic editing} strategy that edits only the image regions corresponding to the question's semantics. Combined with activation patching, this allows us to assess whether the model’s prediction can be causally restored, thereby tracing how semantic signals propagate across layers and modules.

\paragraph{Semantic Editing Prompt Generation.} We construct counterfactual variants of the original question by altering attributes, objects, or relations. We manually design 10 reference examples and use GPT-4o to generate contextually plausible semantic substitutions (e.g., ``red'' to ``green''). To ensure semantic clarity and factual accuracy, we manually verify the generated questions. Based on our verification, less than 1\% cases were filtered out due to abstract descriptions (e.g., “bright” as a color), which are difficult to visualize. Each validated question then serves as a semantic editing prompt, which locates the corresponding image region and applies the intended semantic change. Detailed prompts can be found in Appendix \Scref{sec:prompt:semantic perturbation}.

\paragraph{Semantic Region Editing.} For each question, to enable precise editing of a targeted semantic region while preserving the rest of the image, we require bounding box information for the region of interest, obtained either from the dataset annotations or an off-the-shelf object detection model. We then apply an image segmentation tool, i.e., SAM~\cite{kirillov2023segment}, to segment the target region and use an image editing tool, i.e., PowerPaint~\cite{zhuang2024task}, to locally edit the image, guided by the generated counterfactual prompts, ensuring that only the intended semantics are modified. We manually inspected the visual edits, filtering out $\sim$10\% due to unintended modifications outside the targeted region or unsuccessful edits.

\paragraph{VQA Triple Selection.} To ensure the causal effectiveness of our interventions, we retain only VQA triples where the model correctly predicts the ground-truth answer $y$ on the clean image input $(x,z)$ but fails on the corresponding perturbed input $(x, \tilde{z})$, with $z$ denoting the original image and $\tilde{z}$ the edited one. Specifically, the selected samples must satisfy the following condition:
\begin{equation}
f_\theta(x, z) = y, \quad
f_\theta(x, \tilde{z}) \ne y,
\label{eq:perturb-condition}
\end{equation}
where we compare the model predictions $f_\theta(x, z)$ and $f_\theta(x, \tilde{z})$ with the given correct answer $y$. This filtering criterion ensures that the observed causal effect is both significant and meaningful.

\subsection{Attention Modulating}
\label{sec:seam:attention_masking}

While activation patching is effective for identifying causal layers and token spans, it is limited in granularity and efficiency at the attention head level. First, token-wise patching for each head is computationally expensive. Second, activation-based methods are more suited to identifying helpful structures, but may potentially overlook harmful components that negatively impair visual understanding. To address these issues, we propose a head-level causal \textit{attention modulating} strategy that effectively identifies attention heads critical to visual comprehension, enabling targeted rescaling to enhance model performance. 

\paragraph{Key Attention Head Identification.} The first step is identifying which attention heads are causally responsible for improving or distracting the model's prediction performance. Specifically, let $\mA^l_h\in\sR^{T\times d_h}$ denote the output embedding of the attention head $h\in [1,H]$ from layer $l\in[1,L]$, where $d_h$ is the attention head dimension. We mask an attention head by replacing its output embedding with the average output embedding of the remaining heads at the same layer $l$:

\begin{equation}
\widetilde{\mA}^{l}_{h} = \frac{1}{H - 1} \sum_{h'=1, h' \ne h}^{H} \mA^{l}_{h'}
\label{eq:head-ablation}
\end{equation}

Intuitively, this operation preserves the overall structure of the layer while masking the semantic contribution of head $h$, enabling us to isolate its causal effect. To quantify this effect on a VQA triple $(x,z,y)$, we measure the change in the model's prediction probability of the correct answer token $y$ before and after masking this attention head. Specifically, we pass the masked embedding $\widetilde{\mA}^l_h$ along with the other attention heads' embeddings $\mA_{h'}^l, \forall h'\in [1, H] \wedge h'\ne h$ to the rest of the VLM. This yields a new prediction probability for the correct answer, denoted as $p_\theta(y | x,z; \widetilde{\mA}^l_h)$. For comparison, we also compute the original prediction probability without attention head masking, denoted as $p_\theta(y | x,z)$. The causal effect of head $h$ at layer $l$ is then defined as the change in prediction probability after masking:

\begin{align}
\Delta p_h^l(x, z, y) = p_\theta(y | x,z; \widetilde{\mA}^l_h) - p_\theta(y | x,z)
\label{eq:delta-p}
\end{align}

To identify the positive and negative effects of a head over a dataset $\gD$, we split the dataset into two subsets: the VQA triples predicted by the model correctly and incorrectly, i.e., $\gD_\text{correct}$ and $\gD_\text{incorrect}$. For head $h$ at layer $l$, we compute the average probability change over both subsets:
\begin{align}
    c_\kappa(h,l) = \frac{1}{|\gD_{\kappa}|} \sum_{(x,z,y)\in\gD_{\kappa}} \Delta p_h^l(x,z,y),
\end{align}
where $\kappa=\{\text{correct}, \text{incorrect}\}$. Ranking heads by these scores allows us to identify two sets of heads:

\begin{itemize}[leftmargin=12pt]
  \item \textbf{Positive Heads.} $\gH_\text{pos}$ contains the top-$K$ heads with the highest values of $-c_\text{correct}(h,l)$. Masking these heads leads to the largest drop in prediction probability on $\gD_\text{correct}$, indicating their positive causal effect on correct predictions. 
  \item \textbf{Negative Heads.} $\gH_\text{neg}$ contains the top-$K$ heads with the highest values of $c_\text{incorrect}(h,l)$. Masking these heads leads to the largest gain in prediction probability on $\gD_\text{incorrect}$, indicating their negative causal effect on incorrect predictions.  
\end{itemize}

To align with the three semantic levels, i.e., attributes, objects, and relations, we perform attention head identification separately for each semantic task type, using the corresponding VQA triples. This allows us to identify positive and negative heads that contribute either to general-purpose understanding or to specific semantic categories. 

\begin{figure*}[htbp]
    \centering
    \includegraphics[width=\linewidth]{
    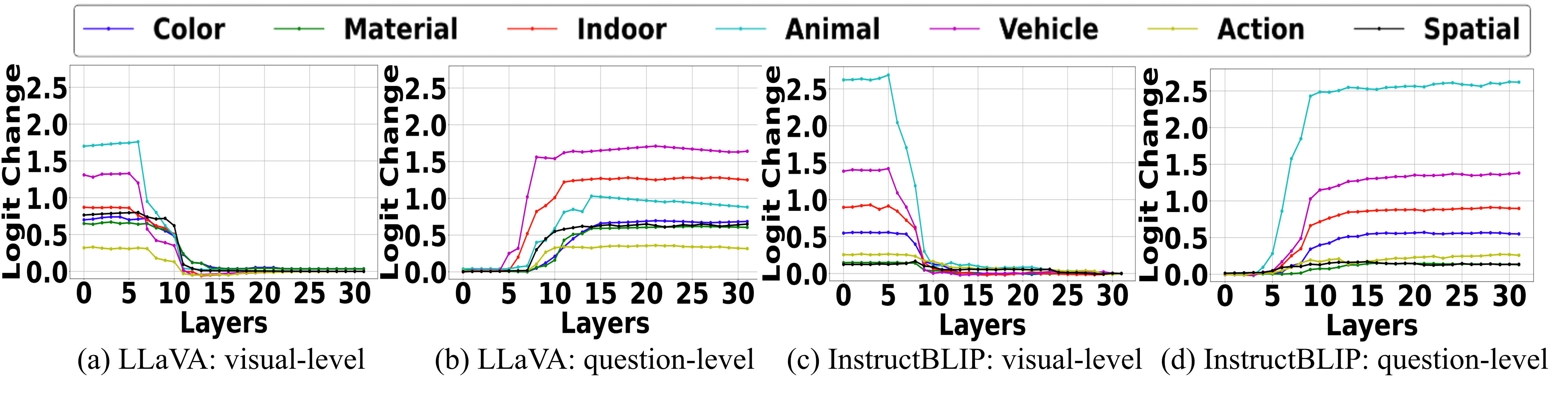}
    \caption{Logit change analysis for image and query token patching in LLaVA and InstructBLIP.}
    \label{fig:logit_changes_image_query_comparison}
    \vspace{-5mm}
\end{figure*}

\paragraph{Attention Head Rescaling.} To verify the causal effect of the heads in $\gH_\text{pos}, \gH_\text{neg}$, we propose an \textit{attention head rescaling} strategy that modulates the influence of these heads to improve visual semantic understanding during inference. 

First, we measure an importance score reflecting the absolute amount of probability changes for the positive and negative heads.
\begin{align}
    c(h,l) = 
    \begin{cases} 
        |c_\text{correct}(h,l)|, & \forall (h, l) \in \gH_\text{pos} \\
        |c_\text{incorrect}(h,l)|, & \forall (h, l) \in \gH_\text{neg}
    \end{cases}    
\end{align}

Next, we normalize the score to a range of $[0,1]$, converting them to a rescaling factor:
\begin{align}
    \lambda(h, l) = \frac{c(h, l) - c_\text{min}}{c_\text{max} - c_\text{min}} \in [0,1],
\end{align}
where $c_\text{min}, c_\text{max}$ are the minimum and maximum values of $c(h,l)$ computed separately for heads in $\gH_\text{pos}$ and $\gH_\text{neg}$. Finally, during inference, we rescale the embeddings $\mA_h^l$ of head $h$ at layer $l$ based on their normalized weights.
\begin{align}
    \widetilde{\mA}_h^l = 
    \begin{cases}
        (1 + \lambda (h, l))\mA_h^l, & (h, l)\in \gH_\text{pos}, \\
        (1 - \lambda (h, l))\mA_h^l, & (h, l)\in \gH_\text{neg}.
    \end{cases}
\end{align}
The key idea is to amplify attention signals from positive heads while suppressing those from negative heads. Importantly, we modify only the outputs of selected heads during the forward pass, leaving the model architecture and parameters unchanged.

\section{VLM Interpretations via \framework}
\label{sec:model_analysis}

 \begin{figure*}[!htbp]
    \centering
    \includegraphics[width=\textwidth]{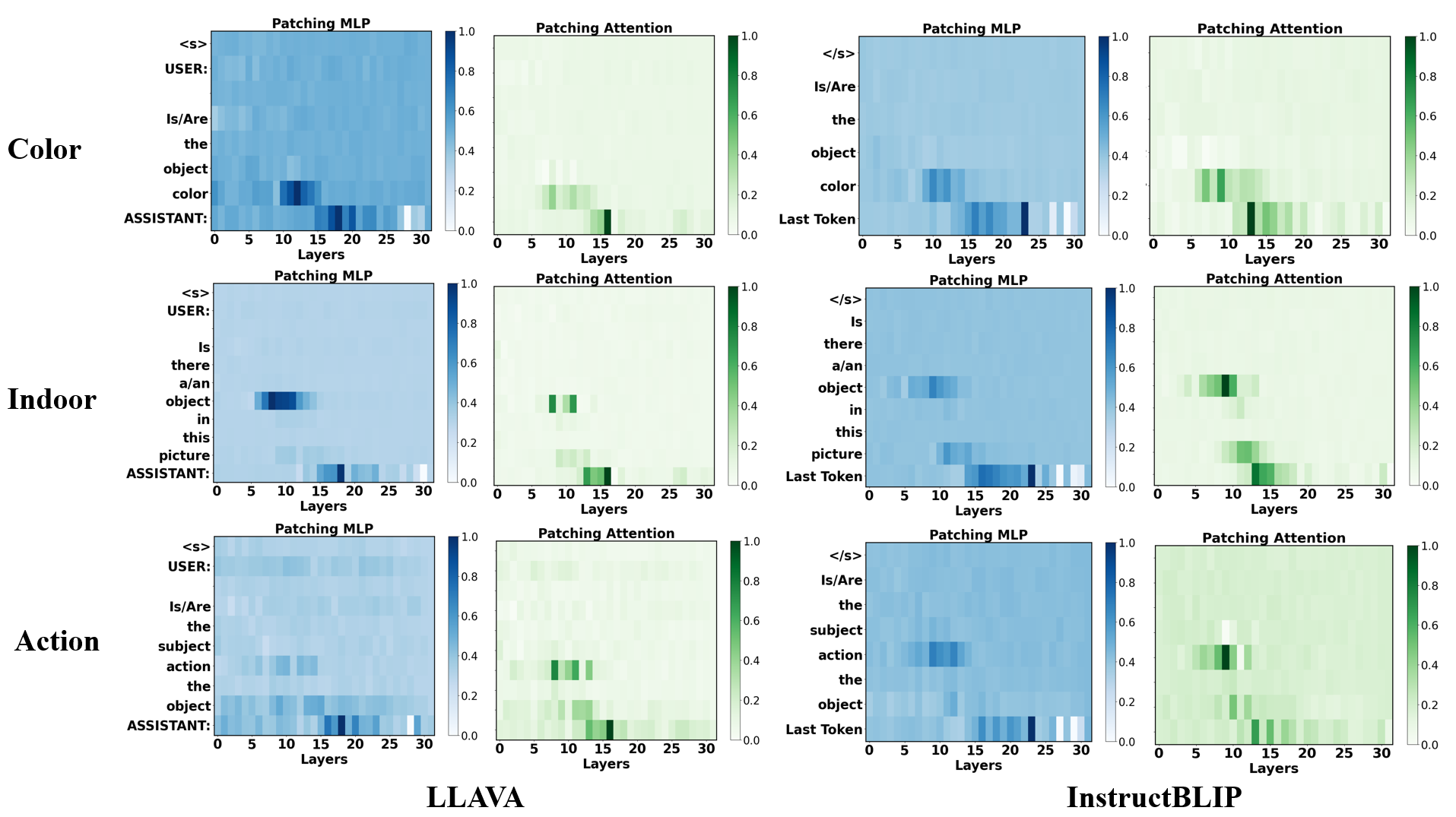}
    \vspace{-3mm}
    \caption{Layer-wise causal impact of MLP (blue) and self-attention (green) on cross-modal semantic understanding. Columns denote transformer layers, rows denote question tokens from a fixed prompt (e.g., “Is the object red?”), and color intensity indicates the $\Delta$logit gain of the correct answer after patching.}
    \label{fig:patch mlp and attention}
    \vspace{-5mm}
\end{figure*}

\subsection{Experimental Setting}
\label{sec:Semantic benchmark}

In this study, we consider both VLM families, with a specific focus on two state-of-the-art VLMs: \textbf{LLaVA 1.5 7B}~\cite{li2024llava} and \textbf{InstructBLIP 7B}~\cite{dai2023instructblip}. Regarding data, we apply \framework using the GQA benchmark \cite{hudson2019gqa}, one of the largest VQA datasets (22M instances), featuring diverse and high-quality annotations of objects, attributes, and relationships across both visual and textual modalities. This makes GQA well-suited for our visual semantic editing to generate corrupted images. Considering this study's scope, we subsample the dataset based on two criteria: (1) we focus on instances involving concepts from high-frequency semantic categories to ensure representativeness; and (2) we prioritize binary, discriminative questions that target a specific semantic property and yield unambiguous answers, reducing data-centric noise in causal tracing analysis. We further balance the distribution of binary question types to ensure equal representation. Following the standard practice of prior work \cite{palit2023towards, golovanevsky2025vlms}, we sample 12,647 questions in total. Table~\ref{tab:dataset_stats} displays the detailed data statistics. Example questions are shown in Appendix~\ref{sec:semantic_benchmark_details}.

\begin{table}[htbp]
\centering
\small
\begin{tabular}{l|l|c}
\toprule
\textbf{Concept-Level} & \textbf{VQA Category} & \textbf{Size} \\
\midrule
Attribute & Material  & 1,300 \\
Attribute & Color     & 1,500 \\
Object    & Animal    & 1,070 \\
Object    & Vehicle   & 1,740 \\
Object    & Indoor    & 2,092 \\
Relation  & Spatial   & 1,950 \\
Relation  & Action    & 2,995 \\
\bottomrule
\end{tabular}
\caption{Dataset sizes for each VQA task categorized by Concept-Level, sorted by size within each Concept.}
\label{tab:dataset_stats}
\end{table}

\subsection{Key Findings}
\label{sec:editing:results}

\paragraph{Multimodal information transfer.} Figure \ref{fig:logit_changes_image_query_comparison} displays the results of information transfer across modalities. Overall, both VLMs start with understanding visual information in the early layers (0-5), then perform visual-text alignment in the middle layers (6–15), and finally shift focus toward the question text for answering in the later layers ($>$15).  Comparatively, text processing requires more layers compared to visual understanding, indicating that VLMs use more parametric memory on the text side. Among semantic categories, both models align object-related information across modalities earlier than relational and attribute information. This pattern is similar to the progress in human visual-language comprehension \cite{ullman1987visual}.

\paragraph{Self-attention and MLP contributions.} 
Given the higher parametric memory demands of text processing compared to visual understanding to answer visual questions, we further analyze key modules (i.e., self-attention and MLP) in the language model within VLMs. Specifically, we apply activation patching~\cite{meng2022locating} to replace corrupted question token embeddings with their clean counterparts at different layers, and measure the logit difference of the correct answer. Figure~\ref{fig:patch mlp and attention} shows the results. 
We observe that both MLP and self-attention store the salient information of target aspect-related tokens (e.g., ``object") in the middle layers (6–12). Looking into the last token, we find that the most influential self-attention layer for answering the question appears earlier (e.g., Layer 16 in LLaVA and Layer 13 in InstructBLIP) compared to the MLP module (e.g., Layer 18 in LLaVA and Layer 23 in InstructBLIP). This suggests that attention contributes more to semantic retrieval at mid-depth, whereas MLPs carry a stronger influence on final decision-making in deeper layers.

To gain deeper insight into the semantic information captured by each module, we project layer-wise self-attention and MLP activations into the vocabulary space using LogitLens~\cite{neo2024towards}. 

\begin{figure*}[htbp]
    \vspace{-4mm}
    \centering
    \includegraphics[width=\linewidth]{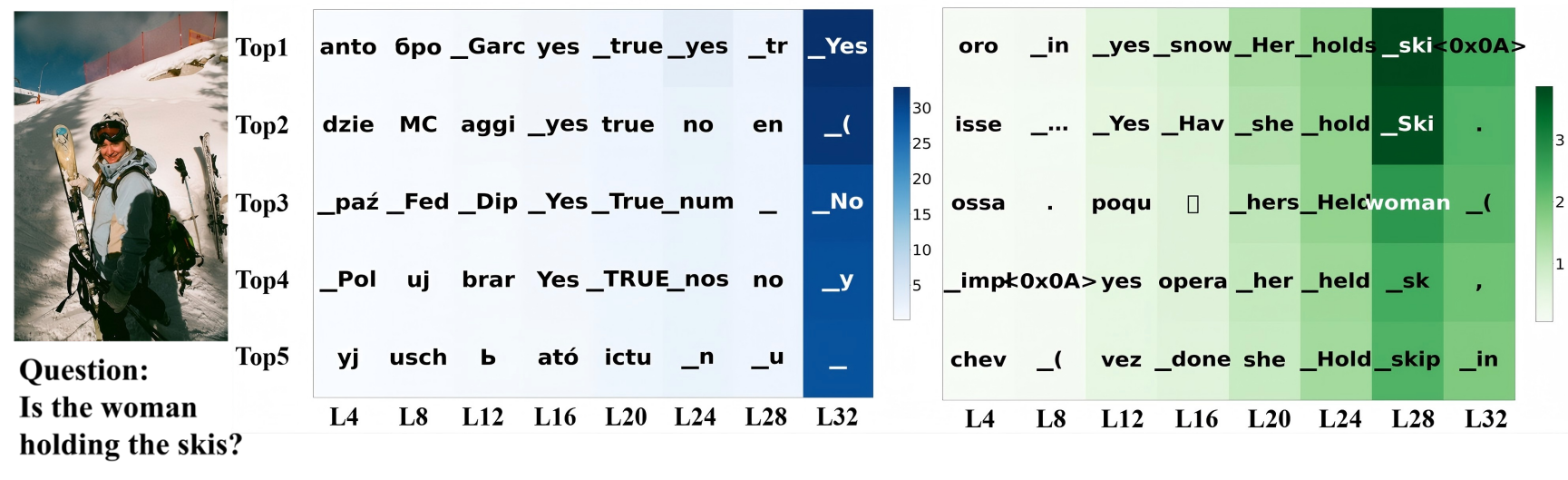}
    \caption{Case study of the MLP (blue) and self-attention (green) logit lens in LLaVA on the Action VQA task. Each row denotes a top-5 predicted token, and each column corresponds to a transformer layer. Color intensity reflects the logit value.}
    \label{fig:logitlens_relation}
    \vspace{-2mm}
\end{figure*}

As shown in Figure~\ref{fig:logitlens_relation}, critical self-attention layers primarily capture answer cues (e.g., ``hold'', ``woman'', and ``ski''), while MLP layers directly focus on the final answer (e.g., ``yes'' or ``no''). Additional examples are shown in Appendix~\ref{sec:logitlens}.

\begin{table*}[htb]
\centering
\small
\renewcommand{\arraystretch}{1.3}
\resizebox{\textwidth}{!}{%
\begin{tabular}{l p{3.2cm} p{3.6cm} p{3.6cm} p{3.6cm}}
\toprule
\multirow{2}{*}{\textbf{Semantic Type}} & 
\multicolumn{2}{c}{\textbf{LLaVA}} & 
\multicolumn{2}{c}{\textbf{InstructBLIP}} \\
\cmidrule(lr){2-3} \cmidrule(lr){4-5}
& \textbf{Positive Heads} & \textbf{Negative Heads} & \textbf{Positive Heads} & \textbf{Negative Heads} \\
\midrule

\textbf{Attribute} & 
{L16.H1,\,L15.H7,\,\newline L22.H11} & 
{L31.H11,\,L30.H11,\newline \,L29.H11} & 
L11.H27, L6.H30 & 
L0.H30, L10.H23, L6.H7, \newline L6.H10, L7.H2, L30.H30\\

\textbf{Object} & 
{L0.H11,\,L26.H11,\,\newline L11.H7} & 
{L31.H11,\,L30.H11,\,L29.H11,\,\newline L12.H2,\,L7.H2,\,L9.H27} & 
L11.H28, L7.H20, \newline L23.H30, L12.H31 & 
L0.H30, L6.H10, L10.H9, \newline L30.H30, L12.H25 \\

\textbf{Relation} & 
{L13.H8,\,L14.H15} & 
{L31.H11,\,L30.H11,\,\newline L29.H11,\,L0.H11} & 
L23.H16, L15.H22, \newline L31.H19, L11.H5, \newline L9.H28, L7.H19, L8.H28 & 
L12.H25, L10.H12, \newline L6.H27, L30.H30 \\

\bottomrule
\end{tabular}%
}
\caption{Shared positive and negative attention heads across semantic types for LLaVA and InstructBLIP. Attention heads are formatted as Layer.Head (e.g., L26.H11).}
\label{tab:head_ablation_table}
\end{table*}

\paragraph{Key attention heads.} Given the role of self-attention layers in identifying answer-relevant evidence, we further identify the key attention heads responsible for capturing this information. Table~\ref{tab:head_ablation_table} lists the most salient heads across semantic levels for both VLMs. Notably, we identify both \textit{positive heads} with stronger causal effects on correct predictions and \textit{negative heads} that introduce more misleading signals compared to randomly selected attention heads. Overall, both types of heads are mutually exclusive, suggesting their distinct roles in VLMs. Interestingly, unlike negative heads that generalize across semantic levels, positive heads are typically shareable within each level but differ across levels. To validate our findings, we measure the spatial alignment between identified heads and object regions. As detailed in Appendix~\ref{sec: semantic attention head finding}, positive heads consistently attend to target-relevant areas, while negative heads often focus on distracting background regions.

\begin{table}[!t]
\centering
\footnotesize
\renewcommand{\arraystretch}{1.1}
\setlength{\tabcolsep}{4pt}
\resizebox{0.48\textwidth}{!}{
\begin{tabular}{llcccc}
\toprule
\textbf{Model} & \textbf{Ablation} & \textbf{Attribute} & \textbf{Object} & \textbf{Relation} & \textbf{Avg} \\
\midrule
\multirow{5}{*}{LLaVA} 
& original & 77.49 & 93.37 & 83.29 & 84.72 \\
& w/o negative & 81.75 & 94.86 & 88.93 & 88.51 \\
& w/o positive & 75.01 & 92.27 & 79.75 & 82.34 \\
& random remove & 77.46 & 93.24 & 83.00 & 84.57 \\
& rescaling & \textbf{82.79} & \textbf{94.99} & \textbf{90.66} & \textbf{89.48} \\
\midrule
\multirow{5}{*}{InstructBLIP} 
& original & 79.46 & 93.12 & 88.55 & 87.04 \\
& w/o negative & 82.78 & 95.24 & 94.52 & 90.85 \\
& w/o positive & 77.85 & 91.08 & 77.84 & 82.26 \\
& random remove & 79.34 & 93.04 & 89.50 & 87.29 \\
& rescaling & \textbf{84.06} & \textbf{95.67} & \textbf{95.21} & \textbf{91.65} \\
\bottomrule
\end{tabular}
}
\caption{Accuracy (\%) of VLMs on VQA under different attention head editing strategies."original" uses all attention heads. "w/o negative" and "w/o positive" remove the top-10 negative or positive heads, respectively. "random remove" randomly drops 10 heads. "rescaling" denotes head embedding rescaling.}
\label{tab:head-edit-comparison}
\vspace{-5mm}
\end{table}

\section{Attention Head Modulating Assessment}
\label{sec:model_editing}



\begin{table}[htbp]
\centering
\renewcommand{\arraystretch}{1.25}
\resizebox{0.48\textwidth}{!}{
\begin{tabular}{lcccccc}
\toprule
\textbf{Category} & \textbf{Initial} & \textbf{10\%} & \textbf{20\%} & \textbf{30\%} & \textbf{50\%} & \textbf{Full} \\
\midrule
Attribute & 77.49 & 79.48 & 81.17 & 82.20 & 82.50 & \textbf{82.79} \\
Object    & 90.18 & 92.42 & 93.20 & 93.24 & \textbf{93.26} & 93.25 \\
Relation  & 83.03 & 87.01 & 90.96 & 91.02 & 91.07 & \textbf{91.11} \\
\midrule
Average   & 83.57 & 86.30 & 88.44 & 88.82 & 88.94 & \textbf{89.05} \\
\bottomrule
\end{tabular}
}
\caption{Accuracy (\%) of \textbf{LLaVA} under different sample proportions for attention head embedding rescaling, grouped by task categories. Bold values indicate best performance for each category.}
\label{tab:soft_reweighting_llava}
\vspace{-3mm}
\end{table}

\paragraph{Effectiveness} We systematically evaluate the impact of editing each identified key attention head on the sampled GQA data, accounting for semantic levels and functional polarity (i.e., positive vs. negative contribution). As shown in Table~\ref{tab:head-edit-comparison}, both VLMs achieve notable performance gains ($\sim$5\% on average) after editing, with the highest improvements on relation-level questions and the lowest on object-level ones. To measure the significance of performance differences between our editing and each baseline, we conduct paired $t$-tests on 1,000 bootstrap-sampled folds per semantic category. As shown in Appendix~\ref{sec:appendix-statistical_significance_analysis}, our rescaling strategy achieves statistically significant improvements ($p < 0.001$) over all four baselines across three semantic levels.

\paragraph{Data Dependence} To assess the method’s data dependence, we randomly sample 10–50\% of the experimental data for key head identification and editing, then test on the full set. Each setting is repeated 10 times to ensure stability, and we report the average results. As shown in Table~\ref{tab:soft_reweighting_llava}, our method achieves comparable performance even with just 10\% of the data, highlighting its applicability in low-resource scenarios.

\paragraph{Generalizability} We further evaluate the generalizability of attention editing on two out-of-distributional (OOD) benchmarks. One is POPE~\cite{li2023evaluating}, which similarly involves binary questions covering three subsets: popular, adversarial, and random. Table~\ref{tab:llava-pope} compares the unedited LLaVA with models edited using in-distribution POPE data (20\%) and GQA data. Both edited versions outperform the unedited model, with the GQA-edited variant even slightly surpassing the one trained on in-distribution data. The other one is COCOQA~\cite{lu2016hierarchical}, which contains open-ended questions spanning color, object, and location categories. We directly evaluate the GQA-edited model against the unedited baseline, and as shown in Appendix~\ref{sec:appendix-attention_editing-results}, both VLMs edited with GQA data continue to outperform the unedited version.

\begin{table}[!t]
\centering
\small
\renewcommand{\arraystretch}{1.15}
\resizebox{0.48\textwidth}{!}{
\begin{tabular}{llccc}
\toprule
\textbf{Task} & \textbf{Metric} & \textbf{Initial} & \textbf{Edit\_POPE} & \textbf{Edit\_GQA} \\
\midrule
\multirow{3}{*}{Popular} 
& Acc     & 0.858 & 0.862 & \textbf{0.866} \\
& Recall  & 0.767 & 0.785 & \textbf{0.798} \\
& F1      & 0.844 & 0.857 & \textbf{0.859} \\
\midrule
\multirow{3}{*}{Adversarial} 
& Acc     & 0.835 & 0.838 & \textbf{0.843} \\
& Recall  & 0.767 & 0.792 & \textbf{0.795} \\
& F1      & 0.823 & \textbf{0.832} & 0.831 \\
\midrule
\multirow{3}{*}{Random} 
& Acc     & 0.867 & 0.880 & \textbf{0.885} \\
& Recall  & 0.767 & 0.795 & \textbf{0.800} \\
& F1      & 0.856 & 0.875 & \textbf{0.878} \\
\bottomrule
\end{tabular}
}
\caption{Performance of LLaVA on POPE under different rescaling strategies. "Initial" refers to the performance without any intervention. "Edit\_POPE" refers to applying head rescaling based on POPE. "Edit\_GQA" denotes using heads and scores discovered on GQA.}
\label{tab:llava-pope}
\vspace{-4mm}
\end{table}

\section{Conclusion}
\label{sec:conclusion}
We present \framework, a novel semantics-based causal interpretability framework for VLMs. We introduce visual semantic editing for concept-level visual interventions and propose attention modulation to identify and edit key attention heads. Through a systematic analysis of LLaVA and InstructBLIP, we find that attention primarily captures salient semantic cues (e.g., color) that guide predictions, while MLP layers are crucial for generating final outputs. We identify top attention heads playing either positive or negative roles. Our attention editing method not only improves model performance but also remains effective in low-resource settings and generalizes well to OOD cases.

\section*{Limitations}
\label{sec:limitations}

Despite the insights presented in this study, several limitations remain:

\begin{enumerate}
    \item To ensure the controllability of causal interventions, we focus primarily on binary discriminative questions with uniform formats (e.g., ``Is the object red?''). While this design facilitates precise patching and attribution, it limits our ability to interpret model behavior in more complex reasoning tasks such as counting, sorting, or multi-hop inference. We leave these aspects for future exploration.
    
    \item Activation patching requires multiple forward passes, especially when intervening across many layers, tokens, or attention heads, which incurs significant computational cost. This limits the scalability of our method to larger models and datasets. Future work may explore more efficient patching strategies to reduce inference time.
    
    \item While we identify reusable ``positive heads'' and ``negative heads'' across various semantic tasks, the mechanisms underlying attention heads in VLMs remain underexplored. For example, it is still unclear whether attention heads interact in cooperative or competitive ways, or whether cross-modal heads exhibit domain-specific dynamic routing. Further analysis is required to systematically understand these behaviors.
\end{enumerate}

\bibliography{custom}

\appendix
\section{Prompt Template}
\label{sec:prompt}

\subsection{Semantic Perturbation Prompts}
\label{sec:prompt:semantic perturbation}

\begin{table*}[htbp]
\centering
\renewcommand{\arraystretch}{1.25}
\resizebox{\textwidth}{!}{
\begin{tabular}{l|l|c|l|l}
\toprule
\textbf{Semantic Type} & \textbf{Original Question} & \textbf{Answer} & \textbf{Full Answer} & \textbf{Counterfactual Question} \\
\midrule
Attribute & Is the shirt \textbf{blue}? & No & The shirt is black. & Is the shirt \textbf{black}? \\
Attribute & Is the chair made of \textbf{wood}? & Yes & The chair is made of wood. & Is the chair made of \textbf{metal}? \\ 
Attribute & Is the car \textbf{black}? & No & The car is white. & Is the car \textbf{white}? \\
Object & Is there a \textbf{dog} in this picture? & No & There is a cat in the image. & Is there a \textbf{cat} in this picture? \\
Object & Is there a \textbf{chair} in this picture? & Yes & There is a chair in the image. & Is there a \textbf{table} in this picture? \\
Object & Is there a \textbf{bus} in this picture? & No & A bicycle is in the image.  & Is there a \textbf{bicycle} in this picture? \\
Relation & Is the person \textbf{riding} the horse? & Yes & The person is riding the horse. & Is the person \textbf{feeding} the horse? \\
Relation & Is the person \textbf{holding} a tennis racket? & Yes & The person is holding a tennis racket. & Is the person \textbf{throwing} a tennis racket? \\
Relation & Is the cat to the \textbf{left} of the sofa? & Yes & The cat is positioned to the left of the sofa. & Is the cat to the \textbf{right} of the sofa? \\
Relation & Is the ball \textbf{under} the table? & No & The ball is on top of the table. & Is the ball \textbf{on top of} the table? \\
\bottomrule
\end{tabular}
}
\caption{Representative examples used for semantic perturbation. Each row shows the original binary question with its answer and full natural-language explanation, followed by a counterfactual version with a modified semantic unit (in bold).}
\label{tab:semantic-perturbation-manualexamples}
\vspace{-4mm}
\end{table*}

To support causal editing in the visual domain, we design semantic perturbation prompts that precisely identify which part of the question's meaning should change. By isolating a single semantic unit—such as an attribute, object, or relation—this strategy provides a grounded and logically consistent basis for local image edits via PowerPaint ~\cite{zhuang2024task}. It also ensures that the intended visual intervention aligns with a meaningful linguistic transformation, enabling controlled and interpretable reasoning analysis.

To automate and accurately generate semantically valid counterfactual questions, we leverage GPT-4o as a language engine. To guide it in producing controlled perturbations, we adopt a few-shot prompting strategy. Specifically, we manually construct 10 reference examples covering three semantic types—attributes, objects, and relations. These examples are used as in-context demonstrations to instruct the model on how to minimally alter a question’s semantic unit while preserving its grammaticality and contextual plausibility.

The prompt consists of three parts: (1) a task-specific instruction, (2) a few in-context examples for the corresponding semantic type, and (3) a new question to be rewritten. In Table~\ref{tab:semantic-perturbation-manualexamples}, we show the complete set of manually created reference examples. These examples serve as the foundation for constructing few-shot prompts, enabling GPT-4o to learn how to generate semantically valid counterfactual questions. Note that in Table~\ref{tab:semantic-perturbation-manualexamples}, the \texttt{Answer} field corresponds to the original binary label provided by the dataset, while the \texttt{Full Answer} is a natural-language explanation included in the GQA annotations to justify the label. We also incorporate them as contextual guidance when generating counterfactual questions, helping to ensure semantic plausibility.

Table~\ref{tab:semantic-perturbation-prompts} provides representative few-shot prompt templates used for each semantic type. The prompts are formatted to clearly separate the original question, the full answer, and the desired counterfactual question. The final line always instructs the model to generate only the rewritten question.

We also applied human filtering to ensure that the model-generated counterfactuals are semantically plausible, grammatically correct, and minimally modified from the original input. All generated questions were manually verified, and less than 1\% were filtered out due to abstract or ambiguous expressions (e.g., ``bright'' as a color) that cannot be reliably grounded in visual content.

\begin{table*}[htbp]
\vspace{-6pt}
\centering
\renewcommand{\arraystretch}{1.3}
\resizebox{\textwidth}{!}{
\begin{tabular}{l|p{0.85\textwidth}}
\toprule
\textbf{Strategy} & \textbf{Examples} \\
\midrule
Attribute Perturbation \\ Few-shot Prompt &
\textbf{You are given a binary Yes/No question about an object's attribute, along with its ground-truth answer and a full explanation.}
Rewrite the question by changing the attribute (e.g., color or material) while preserving the original structure and context.

\textbf{Example 1:}

\texttt{Original Question: Is the shirt blue? Answer: No. Full Answer: The shirt is black. Counterfactual Question: Is the shirt black?}

\textbf{Example 2:}
\texttt{Original Question: Is the chair made of wood? Answer: Yes. Full Answer: The chair is made of wood. Counterfactual Question: Is the chair made of metal?}

\textbf{Example 3:}
\texttt{Original Question: Is the car black? Answer: No. Full Answer: The car is white. Counterfactual Question: Is the car white?}

\textbf{Now please complete the following. Only output the rewritten counterfactual question, without explanation:}
\texttt{Original Question: Is the sofa red? Answer: Yes. Full Answer: The sofa is red. Counterfactual Question:} \\
\midrule
Object Perturbation \\ Few-shot Prompt &
\textbf{You are given a binary Yes/No question about the presence of an object in an image, along with its answer and a full explanation.}
Rewrite the question by changing the queried object to another semantically plausible one.

\textbf{Example 1:}
\texttt{Original Question: Is there a dog in this picture? Answer: No. Full Answer: There is a cat in the image. Counterfactual Question: Is there a cat in this picture?}

\textbf{Example 2:}
\texttt{Original Question: Is there a chair in this picture? Answer: Yes. Full Answer: There is a chair in the image. Counterfactual Question: Is there a table in this picture?}

\textbf{Example 3:}
\texttt{Original Question: Is there a bus in this picture? Answer: No. Full Answer: A bicycle is in the image. Counterfactual Question: Is there a bicycle in this picture?}

\textbf{Now please complete the following. Only output the rewritten counterfactual question, without explanation:}
\texttt{Original Question: Is there a lamp in this picture? Answer: No. Full Answer: There is a shelf in the image. Counterfactual Question:} \\
\midrule
Relation Perturbation \\ Few-shot Prompt &
\textbf{You are given a binary Yes/No question about a spatial or action relationship between objects, along with its ground-truth answer and full explanation.}
Rewrite the question by changing the relation (e.g., left to right, under to above, riding to feeding).

\textbf{Example 1:}
\texttt{Original Question: Is the person riding the horse? Answer: Yes. Full Answer: The person is riding the horse. Counterfactual Question: Is the person feeding the horse?}

\textbf{Example 2:}  
\texttt{Original Question: Is the person holding a tennis racket? Answer: Yes. Full Answer: The person is holding a tennis racket. Counterfactual Question: Is the person throwing a tennis racket?}

\textbf{Example 3:}
\texttt{Original Question: Is the cat to the left of the sofa? Answer: Yes. Full Answer: The cat is positioned to the left of the sofa. Counterfactual Question: Is the cat to the right of the sofa?}

\textbf{Example 4:}
\texttt{Original Question: Is the ball under the table? Answer: No. Full Answer: The ball is on top of the table. Counterfactual Question: Is the ball above the table?}

\textbf{Now please complete the following. Only output the rewritten counterfactual question, without explanation:}
\texttt{Original Question: Is the cup next to the book? Answer: No. Full Answer: The cup is on top of the book. Counterfactual Question:} \\
\bottomrule
\end{tabular}}
\vspace{-8pt}
\caption{Few-shot prompt strategies for semantic perturbation. Each cell lists in-context examples guiding GPT-4o to produce counterfactual binary questions by altering one semantic unit.}
\label{tab:semantic-perturbation-prompts}
\end{table*}

\subsection{Prompt Templates for Evaluation}
\label{sec:prompt:evaluation}

To ensure consistent and interpretable input formatting across evaluation settings, we design task-specific prompt templates for both LLaVA and InstructBLIP. Each template is tailored to the corresponding experimental step, accounting for differences in task format and model instruction style.

For binary judgment tasks—such as activation patching and other classification-based interventions—we adopt a unified yes/no prompt format that explicitly instructs the model to answer concisely. For open-ended tasks like COCOQA, which require free-form answers, we apply lightweight role descriptions and prompt the model to generate concise responses (e.g., single words or phrases). We also align the prompt style with each model’s native instruction tuning paradigm—for instance, LLaVA follows a chat-style interaction between USER and ASSISTANT, while InstructBLIP uses standalone declarative prompts.

Table~\ref{tab:prompt_full_templates} summarizes the full set of prompt templates used in our evaluation pipeline, organized by task type and model.

\begin{table*}[htbp]
\centering
\renewcommand{\arraystretch}{1.25}
\resizebox{\textwidth}{!}{
\begin{tabular}{l|l|l}
\toprule
\textbf{Step} & \textbf{Model} & \textbf{Prompt Template} \\
\midrule
\multirow{2}{*}{Activation Patching (Binary Judgment)} 
& LLaVA & 
\makecell[l]{\textbf{USER:} \\\texttt{<Image>} \\ \texttt{\{Role Description\}}* \\ \texttt{Question: \{Question\}} \\\texttt{Please answer the question using Yes or No.} \\\textbf{ASSISTANT:}} \\
\cmidrule(lr{0.5em}){2-3}
& InstructBLIP & 
\makecell[l]{\texttt{<Image>} \\ \texttt{\{Role Description\}}* \\ \texttt{Question: \{Question\}} \\ \texttt{Please answer the question using yes or no.}} \\
\midrule
\multirow{2}{*}{Closed-form Binary Judgment Tasks (e.g., Color, Material, Indoor, POPE)} 
& LLaVA & 
\makecell[l]{\textbf{USER:} \\\texttt{<Image>} \\ \texttt{\{Role Description\}}* \\ \texttt{Question: \{Question\}} \\\texttt{Please answer the question using Yes or No.} \\\textbf{ASSISTANT:}} \\
\cmidrule(lr{0.5em}){2-3}
& InstructBLIP & 
\makecell[l]{\texttt{<Image>} \\ \texttt{\{Role Description\}}* \\ \texttt{Question: \{Question\}} \\ \texttt{Please answer the question using yes or no.}} \\
\midrule
\multirow{2}{*}{Open-ended Tasks (e.g., COCOQA)} 
& LLaVA & 
\makecell[l]{A chat between a curious user and an artificial intelligence assistant. \\\textbf{USER:} \\\texttt{<Image>} \\ \texttt{\{Role Description\}}* \\ \texttt{Question: \{Question\}} \\ \texttt{Context: N/A} \\ \texttt{Answer the question using a single word or phrase.} \\\textbf{ASSISTANT:}} \\
\cmidrule(lr{0.5em}){2-3}
& InstructBLIP & 
\makecell[l]{\texttt{<Image>} \\ \texttt{\{Role Description\}}* \\ \texttt{Question: \{Question\}} \\ \texttt{Short Answer:}} \\
\bottomrule
\end{tabular}
}
\vspace{-8pt}
\caption{Prompt templates used in different experimental steps for both LLaVA and InstructBLIP. Binary judgment tasks (e.g., activation patching and classification subtasks) adopt yes/no format, while open-ended tasks like COCOQA require concise short-form answers.}
\label{tab:prompt_full_templates}
\end{table*}

\section{Semantic Benchmark Details}
\label{sec:semantic_benchmark_details}

To ensure the feasibility of controlled causal intervention and minimize semantic ambiguity, we construct fine-grained subsets from the GQA dataset based on the following principles:

\begin{itemize}
    \item \textbf{Question Type Filtering:} We retain only binary, fact-based questions with a uniform format and clear Yes/No answers. These typically involve single entities and single predicates, such as “Is the object red?” or “Is there a dog in this picture?”.
    
    \item \textbf{High-Frequency Semantics:} Within each semantic category—attributes, objects, and relations—we select commonly occurring subtypes to ensure sufficient coverage, reduce sparsity, and support statistically robust evaluation.

    \item \textbf{Structural Consistency:} We retain only syntactically simple and semantically clear questions, such as those with a single subject–verb–object structure. Questions involving three or more entities or compound constructions are excluded to ensure consistency in question type and support reliable semantic interventions.

    \item \textbf{Answer Balance:} Each semantic subset is curated to maintain a roughly equal distribution of \texttt{Yes} and \texttt{No} answers (approximately 50\% each), thereby reducing label bias and enabling fairer evaluation of model behavior under semantic perturbation.
    
    \item \textbf{Benchmark Representativeness:} Our subset selection process is informed by prior interpretability work~\cite{palit2023towards, golovanevsky2025vlms}. Moreover, the sample size of each subset is comparable to or larger than those in related studies.
\end{itemize}

Based on the above selection principles, we organize the curated VQA questions into a three-level semantic hierarchy—attributes, objects, and relations—each capturing commonly encountered visual concepts in real-world images.

To enhance semantic coverage and ensure representativeness, we select seven high-frequency subtypes across the three semantic levels: for attributes, we include color and material; for objects, we include indoor items, animals, and vehicles; and for relations, we cover spatial and action relations. These subtypes span a wide range of visual semantics and support diverse yet interpretable evaluation.

Table~\ref{tab:semantic_stats} provides an overview of these subtypes along with representative question examples.

\begin{table*}[ht]
\centering
\small
\begin{tabular}{|>{\centering\arraybackslash}m{2.8cm}
                |>{\centering\arraybackslash}m{4.5cm}
                |>{\centering\arraybackslash}m{2.2cm}
                |>{\centering\arraybackslash}m{4.4cm}|}
\toprule
\textbf{Semantic Type} & \textbf{Example Question} & \textbf{Samples} & \textbf{Image Example} \\
\midrule
\textbf{Attribute} & & & \\
\quad Material & Is the chair made of Wood? & 1,300 & \includegraphics[height=1.2cm]{Appendix/material.jpg} \\
\quad Color & Is the shirt blue? & 1,500 & \includegraphics[height=1.2cm]{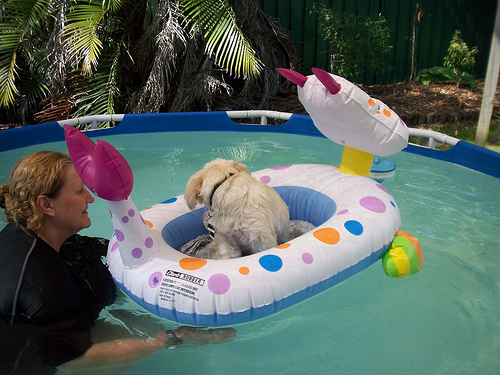} \\
\midrule
\textbf{Object} & & & \\
\quad Animal & Is there a dog in this picture? & 1,070 & \includegraphics[height=1.2cm]{Appendix/animal.jpg} \\
\quad Vehicle & Is there a bike in this picture? & 1,740 & \includegraphics[height=1.2cm]{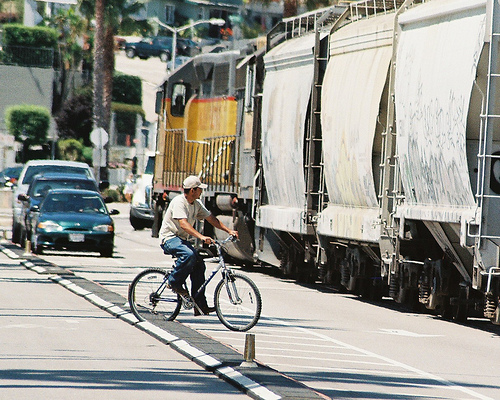} \\
\quad Indoor & Is there a mirror in this picture? & 2,092 & \includegraphics[height=1.2cm]{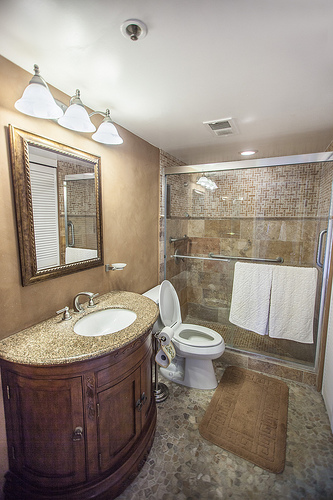} \\
\midrule
\textbf{Relation} & & & \\
\quad Spatial & Is the man right of the horse? & 1,950 & \includegraphics[height=1.2cm]{Appendix/spatial.jpg} \\
\quad Action & Is the woman holding the skis? & 2,995 & \includegraphics[height=1.2cm]{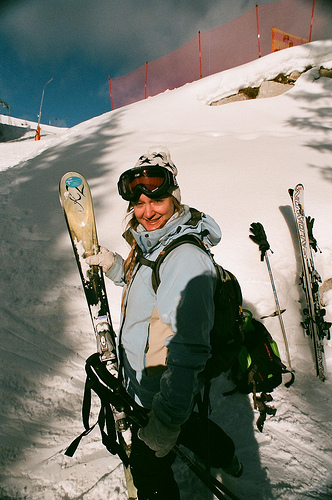} \\
\bottomrule
\end{tabular}
\caption{Statistics and example images for each VQA task in our semantic benchmark. Semantic subtypes are grouped under attribute, object, and relation categories.}
\label{tab:semantic_stats}
\end{table*}

\section{Comparison of Perturbation Methods}
\label{sec:appendix-perturbation-comparison}

\begin{figure*}[ht]
    \centering
    \includegraphics[width=\textwidth]{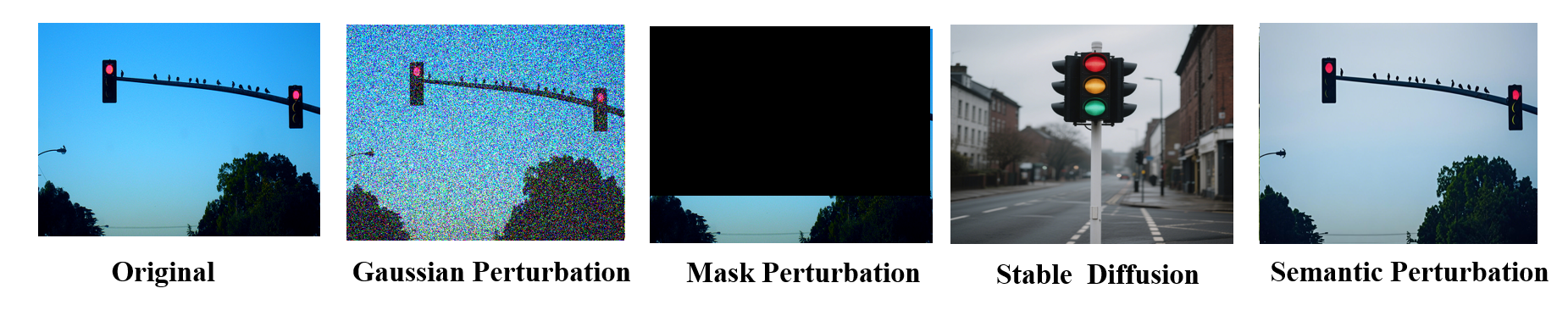}
    \caption{Visual comparison of different perturbation strategies for the object-level task ("Is the sky blue?"). Our semantic perturbation method (right) maintains global visual coherence while precisely intervening on the targeted semantic region (the sky). Traditional methods, including Gaussian noise, regional masking, and full-image redrawing with Stable Diffusion, introduce larger visual distortions and unintended semantic changes.}
    \label{fig:perturbation_visual_attribute}
\end{figure*}

\begin{figure*}[ht]
    \centering
    \includegraphics[width=\textwidth]{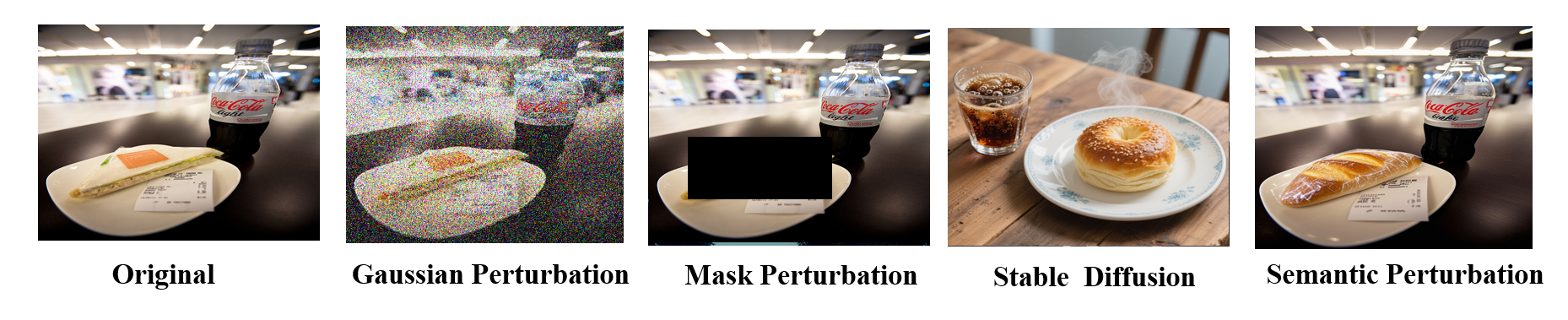}
    \caption{Visual comparison of different perturbation strategies for the object-level task ("Is there a sandwich in this picture?"). Our semantic perturbation method (right) maintains global visual coherence while precisely modifying the presence of the target object. Traditional methods, including Gaussian noise, regional masking, and full-image redrawing with Stable Diffusion, introduce significant visual distortion and lack fine-grained semantic control.}
    \label{fig:perturbation_visual_object}
\end{figure*}

\begin{figure*}[ht]
    \centering
    \includegraphics[width=\textwidth]{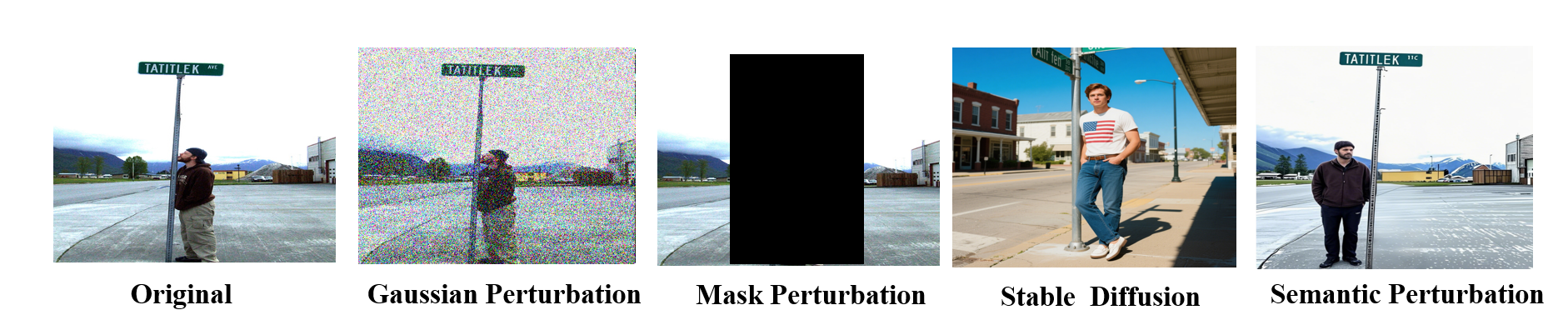}
    \caption{Visual comparison of different perturbation strategies for the relation-level task ("Is the man standing to the left of the signpost?"). Our semantic perturbation method (right) maintains global visual coherence while precisely modifying the spatial relationship between the target entities (man and signpost). In contrast, traditional approaches—such as Gaussian noise, regional masking, and full-image redrawing via Stable Diffusion—introduce substantial visual artifacts and fail to offer controlled, semantics-aware editing.}
    \label{fig:perturbation_visual_relation}
\end{figure*}

To validate the effectiveness of our proposed semantic perturbation method in preserving the global semantic consistency of the input image, we compare it against several commonly used perturbation strategies, including Gaussian noise, salt-and-pepper noise, masking, and diffusion-based image editing. Traditional methods typically apply random noise or local occlusion to the image, which often disrupts low-level features and structural coherence, thereby distorting the distribution of visual features extracted by the encoder. In contrast, our method leverages scene graphs and the SAM model, combined with state-of-the-art image editing techniques, to precisely modify only the targeted semantic region while leaving the rest of the image intact. As illustrated in Figures~\ref{fig:perturbation_visual_attribute}, \ref{fig:perturbation_visual_object}, and~\ref{fig:perturbation_visual_relation}, our semantic perturbation approach maintains global visual coherence while achieving fine-grained, interpretable edits across attribute-level, object-level, and relation-level tasks, respectively.

To quantitatively assess the distributional shift introduced by different perturbation methods, we extract global image features from both the original and perturbed images using the visual encoder employed by LLaVA (CLIP-ViT-L), and compute the cosine similarity between them. Theoretically, if a perturbation only modifies a localized semantic element, the global feature representation should remain highly similar to the original. In contrast, traditional perturbations such as noise or masking often cause significant degradation in this similarity. Tables~\ref{tab:perturbation_similarity_attribute}, \ref{tab:perturbation_similarity_object}, and~\ref{tab:perturbation_similarity_action} report the average cosine similarity across three representative semantic tasks. Our method consistently yields the highest similarity, indicating minimal disturbance to global representations.

Experimental results demonstrate that our method consistently yields higher cosine similarity across all semantic tasks, indicating that it achieves effective semantic intervention with minimal disturbance to the global image distribution. In contrast, Gaussian noise, salt-and-pepper noise, masking, and full-image redrawing via diffusion models introduce substantial low-level feature distortions, leading to more pronounced shifts in the visual representation space.

\begin{table}[ht]
\centering
\small
\captionsetup{position=bottom}
\begin{tabular}{l c}
\toprule
\textbf{Perturbation Method} & \textbf{Cosine Similarity} \\
\midrule
Gaussian Noise – Intensity 10 & 0.8766 \\
Gaussian Noise – Intensity 30 & 0.6333 \\
Gaussian Noise – Intensity 50 & 0.5025 \\
Gaussian Noise – Intensity 80 & 0.3446 \\
Salt-and-Pepper Noise – Intensity 10 & 0.7088 \\
Salt-and-Pepper Noise – Intensity 30 & 0.5554 \\
Salt-and-Pepper Noise – Intensity 50 & 0.4110 \\
Salt-and-Pepper Noise – Intensity 80 & 0.3142 \\
Masking & 0.6005 \\
Diffusion-Based Redrawing & 0.6891 \\
\textbf{Semantic Perturbation (Ours)} & \textbf{0.9168} \\
\bottomrule
\end{tabular}
\caption{Cosine similarity between original and perturbed image features under different perturbation strategies. Feature representations are extracted from the vision encoder of the LLaVA model (CLIP-ViT-L) on the Color VQA dataset. Higher similarity indicates better preservation of global visual semantics, which is crucial for stable reasoning in real-world environments.}
\label{tab:perturbation_similarity_attribute}
\end{table}

\begin{table}[ht]
\centering
\small
\begin{tabular}{l c}
\toprule
\textbf{Perturbation Method} & \textbf{Cosine Similarity} \\
\midrule
Gaussian Noise – Intensity 10 & 0.8628 \\
Gaussian Noise – Intensity 30 & 0.8148 \\
Gaussian Noise – Intensity 50 & 0.7206 \\
Gaussian Noise – Intensity 80 & 0.5988 \\
Salt-and-Pepper Noise – Intensity 10 & 0.7427 \\
Salt-and-Pepper Noise – Intensity 30 & 0.6729 \\
Salt-and-Pepper Noise – Intensity 50 & 0.6159 \\
Salt-and-Pepper Noise – Intensity 80 & 0.5840 \\
Masking & 0.6821 \\
Diffusion-Based Redrawing & 0.6205 \\
\textbf{Semantic Perturbation (Ours)} & \textbf{0.8970} \\
\bottomrule
\end{tabular}
\caption{Cosine similarity between original and perturbed image features under different perturbation strategies. Feature representations are extracted from the vision encoder of the LLaVA model (CLIP-ViT-L) on the Indoor VQA dataset. Higher similarity indicates better preservation of global visual semantics, which is crucial for stable reasoning in real-world environments.}
\label{tab:perturbation_similarity_object}
\end{table}

\begin{table}[ht]
\centering
\small
\begin{tabular}{l c}
\toprule
\textbf{Perturbation Method} & \textbf{Cosine Similarity} \\
\midrule
Gaussian Noise – Intensity 10 & 0.7921 \\
Gaussian Noise – Intensity 30 & 0.7430 \\
Gaussian Noise – Intensity 50 & 0.6204 \\
Gaussian Noise – Intensity 80 & 0.4877 \\
Salt-and-Pepper Noise – Intensity 10 & 0.7033 \\
Salt-and-Pepper Noise – Intensity 30 & 0.6182 \\
Salt-and-Pepper Noise – Intensity 50 & 0.5820 \\
Salt-and-Pepper Noise – Intensity 80 & 0.5194 \\
Masking & 0.4213 \\
Diffusion-Based Redrawing & 0.5586 \\
\textbf{Semantic Perturbation (Ours)} & \textbf{0.8542} \\
\bottomrule
\end{tabular}
\caption{Cosine similarity between original and perturbed image features under different perturbation strategies. Feature representations are extracted from the vision encoder of the LLaVA model (CLIP-ViT-L) on the Action VQA dataset. Results show lower similarity compared to object-level tasks, reflecting higher sensitivity of action semantics to visual changes.}
\label{tab:perturbation_similarity_action}
\end{table}

\section{Logit Lens Analysis of MLP and Attention Projections}
\label{sec:logitlens}

To better understand the reasoning process within vision-language models, we apply the Logit Lens method to analyze the token prediction distributions across layers in the language models of LLAVA and InstructBLIP. Following the standard procedure described by \cite{neo2024towards}, we compare the output logits projected from the MLP pathway and the attention output pathway at each transformer layer.

In  VQA tasks (e.g., “Is the chair made of wood?”), we observe that in the early layers (Layers 2–7), both MLP and attention projections primarily output low-level or semantically ambiguous tokens, such as punctuation, garbled text, or subword fragments, indicating that meaningful semantic representations have not yet formed. Around Layer 9, the attention projection begins to produce answer-relevant tokens (e.g., \texttt{yes}), marking the onset of vision-language alignment. In contrast, the MLP projection only begins to consistently generate high-confidence answer tokens (e.g., \texttt{yes}, \texttt{true}, \texttt{number}) after Layer 15.

Between Layers 16–25, the attention projections increasingly focus on image-grounded semantic concepts (e.g., \texttt{wood}, \texttt{material}, \texttt{color}), exhibiting strong semantic sensitivity. Meanwhile, the MLP projections focus more on decision-related tokens, with steadily increasing confidence in output logits. In the final layer, the MLP projection directly outputs the final answer token with high certainty, while the attention projection tends to focus on formatting or structural tokens (e.g., \texttt{<0x0A>}, \texttt{.}), reflecting its role in output structuring.

\begin{figure*}[htbp]
    \centering
    \includegraphics[width=\linewidth]{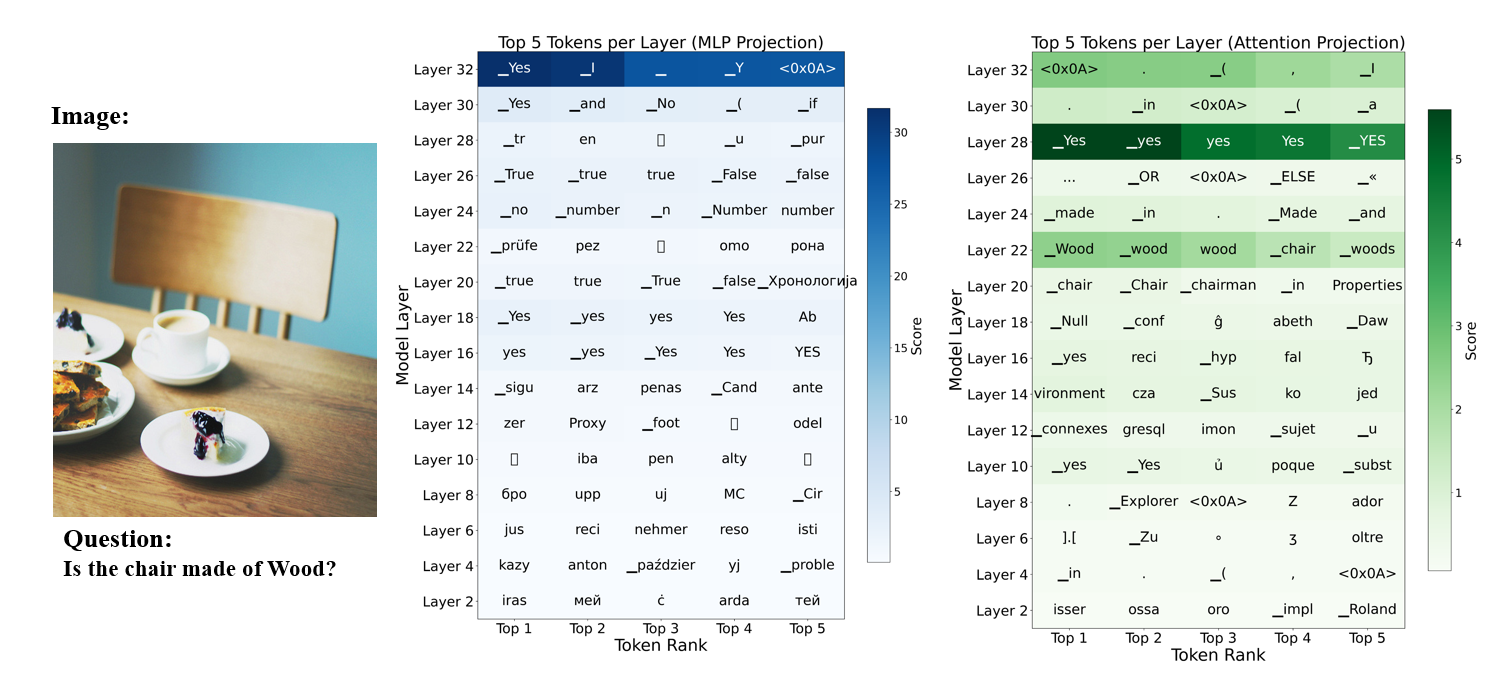}
    \caption{Case of  MLP (blue) and self-attention (green) Logit Lens in LLAVA on Material VQA.}
    \label{fig:logitlens_attribute}
\end{figure*}

\begin{figure*}[htbp]
    \centering
    \includegraphics[width=\linewidth]{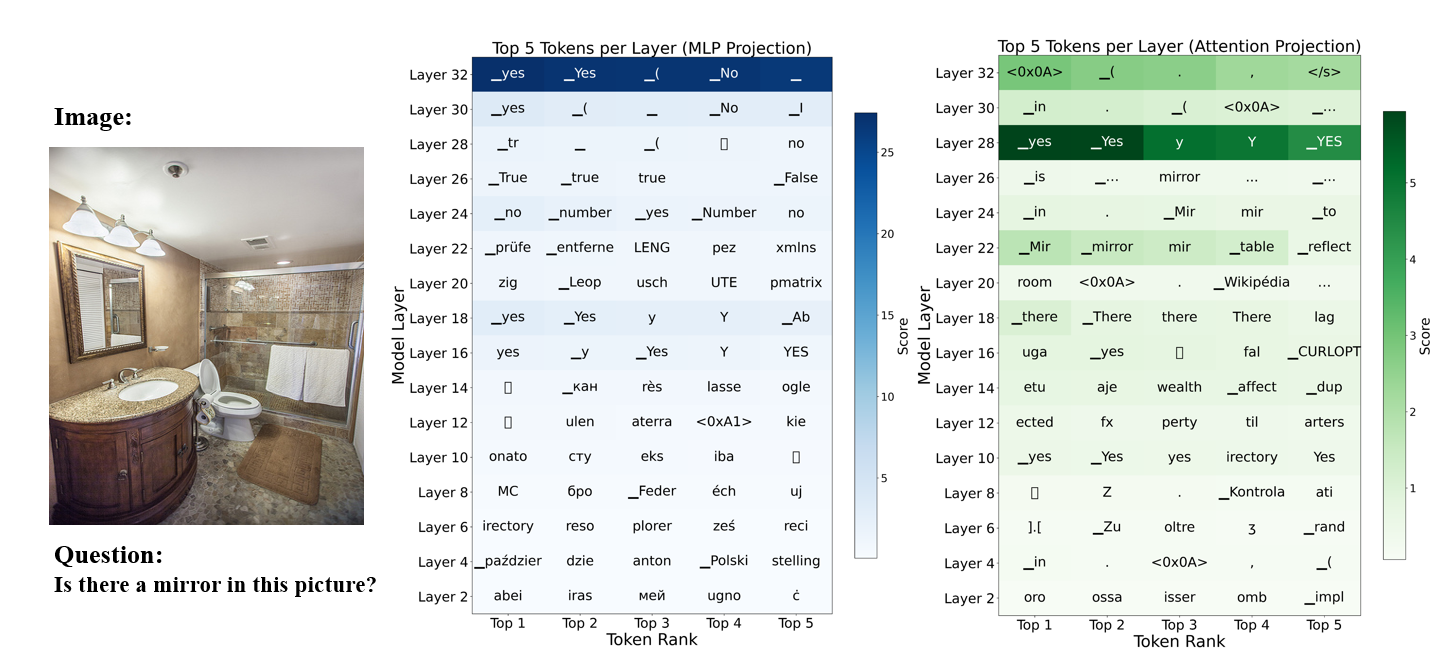}
    \caption{Case of  MLP (blue) and self-attention (green) Logit Lens in InstructBLIP on Indoor VQA.}
    \label{fig:logitlens_object}
\end{figure*}

These findings support a stage-wise interpretation of the hierarchical reasoning process in VQA. Early layers are mainly involved in visual signal processing, the middle layers (Layers 8–16) are responsible for cross-modal alignment and semantic grounding, and the deeper layers (Layers 16–32) integrate language reasoning and answer generation. Our Logit Lens heatmap analysis (Figure~\ref{fig:logitlens_attribute} and ~\ref{fig:logitlens_object}) clearly illustrates this semantic progression: predicted tokens evolve from noisy or ambiguous outputs to aligned concepts and finally converge on task-specific, high-confidence answers. Moreover, the analysis reveals a clear division of labor across layers: attention mechanisms are primarily responsible for multimodal alignment and semantic construction, while MLP projections focus on decision-making.

\section{Activation Patching Supplementary Results}
\label{sec:activation_patching_results}

\begin{figure*}[htbp]
    \centering
    \includegraphics[width=\textwidth]{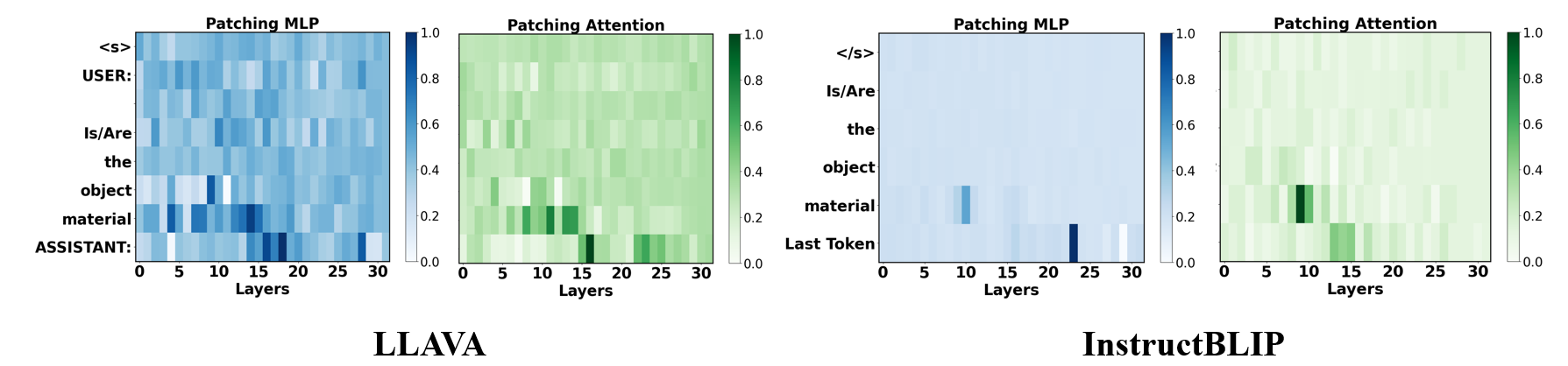}  
    \caption{Layer-wise causal impact of MLP and self-attention on the Material VQA task}
    \label{fig: Material patch mlp and attention}
\end{figure*}

\begin{figure*}[htbp]
    \centering
    \includegraphics[width=\textwidth]{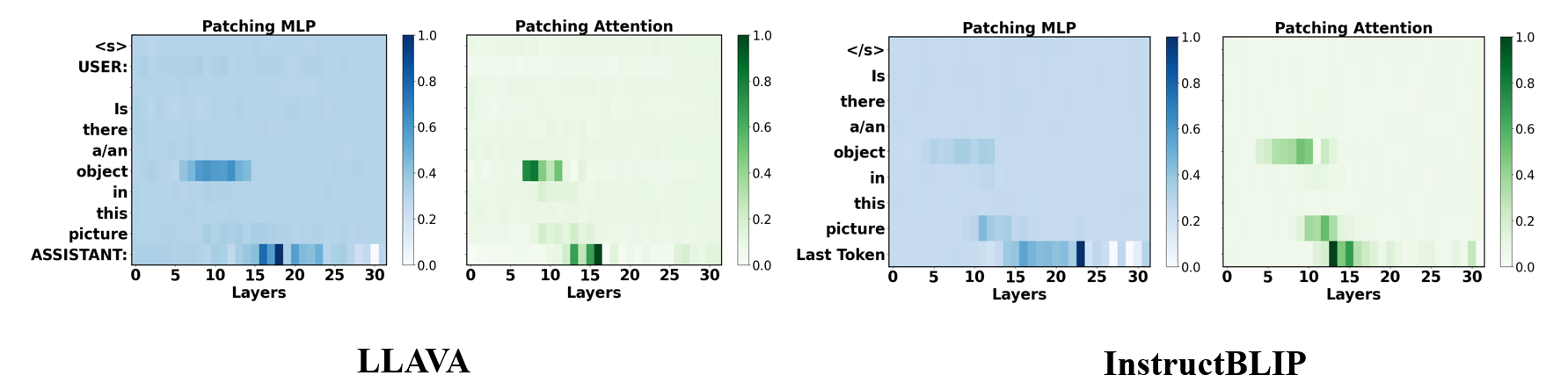}  
    \caption{Layer-wise causal impact of attention and MLP on the Vehicle VQA task}
    \label{fig: Vehicle patch mlp and attention}
\end{figure*}

\begin{figure*}[htbp]
    \centering
    \includegraphics[width=\textwidth]{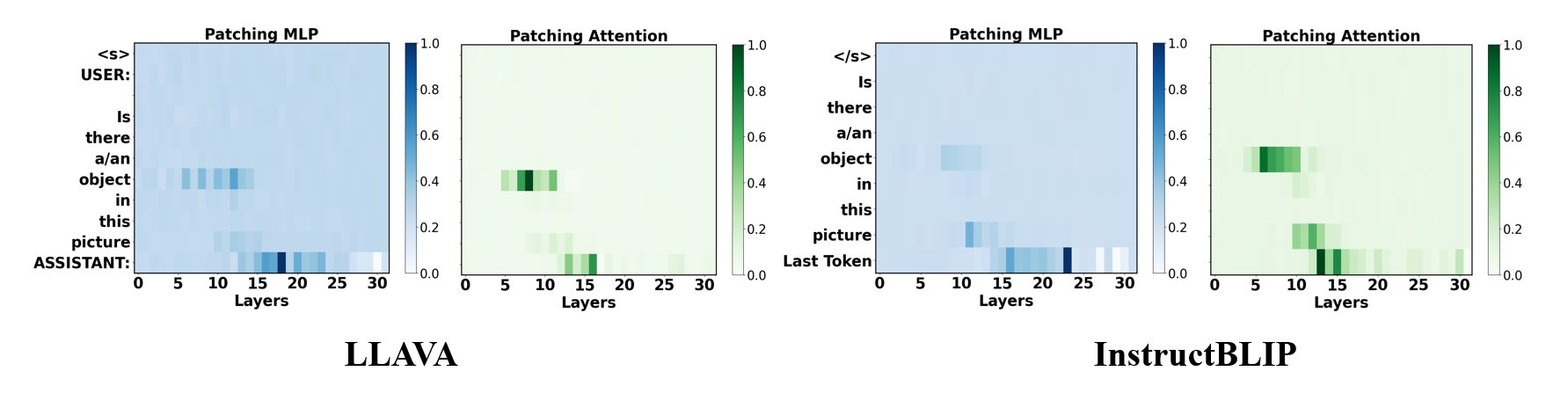}  
    \caption{Layer-wise causal impact of MLP and self-attention on the Animal VQA task}
    \label{fig: Animal patch mlp and attention}
\end{figure*}

\begin{figure*}[htbp]
    \centering
    \includegraphics[width=\textwidth]{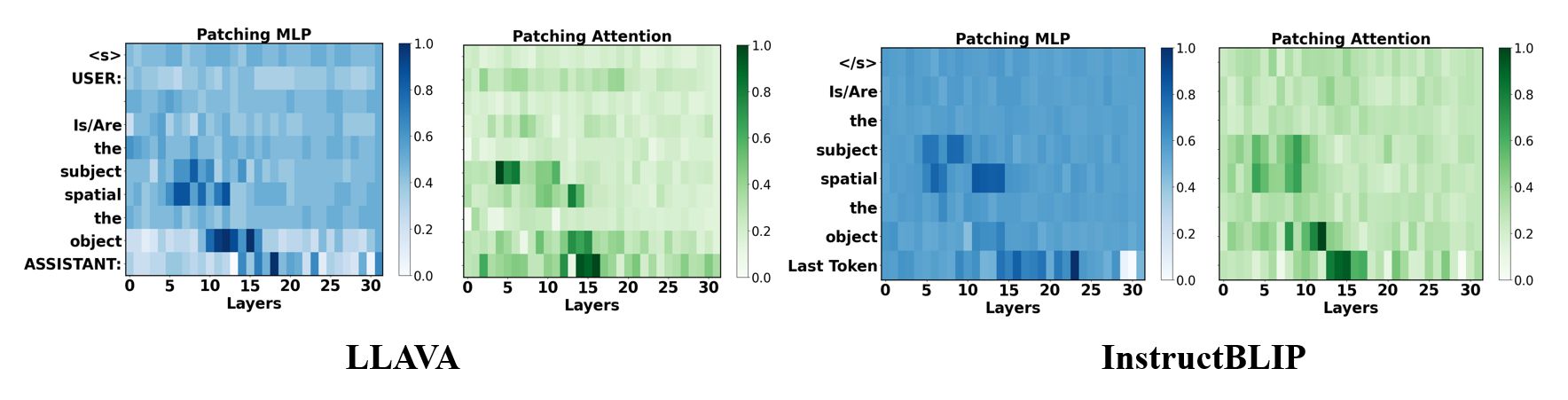}  
    \caption{Layer-wise causal impact of MLP and self-attention on the Spatial VQA task}
    \label{fig: Spatial patch mlp and attention}
\end{figure*}

In this section, we supplement our analysis with activation patching experiments on four VQA sub-tasks—\textbf{Material}, \textbf{Vehicle}, \textbf{Animal}, and \textbf{Spatial}—to systematically examine the causal roles of the language model’s \textbf{MLP} and \textbf{Attention} layers in the information flow. As shown in Figures~\ref{fig: Material patch mlp and attention}, \ref{fig: Vehicle patch mlp and attention}, \ref{fig: Animal patch mlp and attention}, and \ref{fig: Spatial patch mlp and attention}. These experiments are conducted on two representative multimodal models, \textbf{LLaVA} and \textbf{InstructBLIP}, both of which exhibit highly consistent causal response patterns and functional specialization across different semantic tasks. These findings further support our main analysis regarding the separation of reasoning phases and the division of roles between MLP and Attention layers in multimodal reasoning.

\begin{table*}[htbp]
\centering
\scriptsize
\renewcommand{\arraystretch}{1.15}
\begin{tabular}{p{2.8cm} p{6cm} p{6cm}}
\toprule
\textbf{VQA Task (Dataset)} & \textbf{LLAVA (Positive/Negative Heads)} & \textbf{InstructBLIP (Positive/Negative Heads)} \\
\midrule

Color Attribute Positive &
L6 H19, L26 H15, L16 H4, L11 H29, L15 H7, L16 H1, L0 H11, L22 H11, L7 H23, L24 H11 &
L11 H26, L10 H7, L6 H30, L9 H17, L10 H9, L6 H10, L10 H23, L6 H27, L11 H27, L11 H11 \\

Color Attribute Negative &
L31 H11, L30 H11, L29 H11, L9 H12, L26 H11, L7 H27, L15 H26, L10 H18, L10 H1, L15 H3 &
L0 H30, L10 H23, L6 H7, L29 H30, L11 H26, L6 H10, L10 H7, L7 H2, L30 H30, L9 H18 \\

Material Attribute Positive &
L12 H23, L16 H1, L3 H14, L13 H8, L6 H17, L9 H22, L14 H18, L22 H11, L8 H15, L15 H7 &
L23 H30, L7 H5, L6 H30, L11 H5, L8 H21, L7 H19, L11 H27, L5 H28, L11 H20, L12 H6 \\

Material Attribute Negative &
L30 H11, L31 H11, L29 H11, L11 H29, L3 H11, L2 H11, L4 H11, L9 H24, L10 H16, L7 H12 &
L0 H30, L10 H23, L6 H10, L7 H2, L30 H30, L10 H12, L13 H19, L11 H27, L6 H7, L15 H5 \\

Indoor Object Positive &
L0 H11, L0 H23, L26 H11, L7 H20, L2 H16, L1 H17, L14 H18, L6 H26, L22 H11, L11 H7 &
L23 H30, L11 H23, L7 H20, L11 H28, L23 H30, L9 H0, L12 H31, L6 H29, L11 H6, L7 H19 \\

Indoor Object Negative &
L31 H11, L30 H11, L29 H11, L12 H2, L7 H2, L2 H11, L10 H1, L9 H27, L12 H29, L12 H13 &
L0 H30, L8 H27, L6 H10, L30 H30, L12 H25, L10 H9, L13 H19, L14 H25, L30 H4, L10 H8 \\

Animal Object Positive &
L0 H11, L26 H11, L1 H2, L11 H3, L11 H7, L2 H2, L6 H8, L15 H11, L10 H26, L11 H1 &
L23 H30, L5 H28, L23 H30, L9 H0, L11 H28, L15 H22, L7 H20, L7 H13, L7 H11, L12 H31 \\

Animal Object Negative &
L31 H11, L30 H11, L12 H2, L29 H11, L2 H31, L7 H2, L9 H27, L7 H9, L5 H0, L7 H1 &
L0 H30, L6 H10, L10 H9, L0 H28, L12 H25, L13 H28, L10 H12, L30 H30, L6 H7, L5 H16 \\

Vehicle Object Positive &
L0 H11, L26 H11, L8 H11, L2 H2, L1 H2, L11 H7, L10 H15, L11 H17, L13 H4, L11 H1 &
L23 H30, L7 H20, L11 H28, L12 H22, L11 H5, L12 H31, L10 H10, L8 H13, L9 H28, L31 H19 \\

Vehicle Object Negative &
L31 H11, L30 H11, L29 H11, L12 H2, L9 H30, L11 H30, L9 H27, L12 H29, L7 H2, L12 H13 &
L0 H30, L6 H10, L10 H9, L30 H30, L12 H25, L7 H4, L6 H7, L6 H27, L8 H27, L10 H12 \\

Action Relation Positive &
L16 H1, L26 H11, L1 H16, L13 H8, L8 H13, L8 H10, L12 H16, L14 H15, L5 H16, L13 H24 &
L23 H16, L3 H30, L8 H28, L15 H22, L10 H11, L11 H5, L9 H28, L7 H19, L10 H6, L31 H19 \\

Action Relation Negative &
L31 H11, L30 H11, L29 H11, L12 H14, L10 H1, L1 H4, L12 H2, L5 H19, L0 H11, L2 H11 &
L6 H27, L12 H25, L10 H9, L7 H3, L6 H10, L10 H0, L10 H12, L30 H30, L30 H4, L7 H27 \\

Spatial Relation Positive &
L10 H17, L14 H15, L15 H18, L18 H17, L10 H14, L8 H27, L9 H28, L2 H23, L13 H8, L8 H20 &
L23 H16, L15 H22, L31 H19, L11 H5, L11 H28, L10 H1, L7 H19, L9 H28, L5 H30, L8 H28 \\

Spatial Relation Negative &
L31 H11, L30 H11, L29 H11, L0 H11, L15 H23, L15 H10, L15 H4, L3 H11, L6 H14, L1 H16 &
L10 H10, L12 H25, L6 H27, L13 H19, L0 H30, L12 H27, L5 H23, L30 H30, L10 H12 \\
\bottomrule
\end{tabular}
\vspace{-1em}
\caption{Identified Top 10 positive and negative attention heads for different semantic VQA tasks in LLAVA and InstructBLIP.}
\label{tab:appendix-attention-heads}
\end{table*}

\begin{table}[h]
\centering
\small
\begin{tabular}{lccc}
\toprule
\textbf{Dataset} & \textbf{Head Group} & \textbf{LLaVA} & \textbf{InstructBLIP} \\
\midrule
\multirow{2}{*}{Animal} & Positive & 83.8 & 79.2 \\
                        & Negative & 7.1 & 8.5 \\
\midrule
\multirow{2}{*}{Action} & Positive & 87.0 & 85.4 \\
                        & Negative & 15.6 & 13.9 \\
\bottomrule
\end{tabular}
\vspace{0.5em}
\caption{Average Overlap (\%) of Attention Inside Object Bounding Boxes. Higher values indicate stronger alignment with target objects. Results are computed over the top-10 positive/negative heads per model and averaged across samples.}
\label{tab:attn_overlap}
\end{table}

\section{Semantic Attention Head Supplementary Results}
\label{sec: semantic attention head finding}

\paragraph{Comprehensive Attention Head Attribution Results by Semantic Category.}
To complement the main paper's analysis of shared attention heads across semantic levels, we provide the complete attribution results for each VQA task in Table~\ref{tab:appendix-attention-heads}. For every semantic subtype (e.g., color, material, spatial relation), we identify the Top-10 \textit{reasoning heads} and Top-10 \textit{distraction heads} based on their estimated causal contributions to the model’s output, as measured by attention-head ablation and activation patching methods described in Section~\ref{sec:interpretation_method}.

We list the most influential heads in both LLaVA and InstructBLIP, denoted as \texttt{L\# H\#}, where \texttt{L} refers to the transformer layer index and \texttt{H} to the head index within the layer. 

Notably, several trends are consistent across models and semantic categories:
\begin{itemize}
    \item In LLAVA, Positive heads tend to cluster in the mid-to-late layers (e.g., Layers 10–23), aligning with the model’s transition from perceptual alignment to abstract reasoning.
    \item In LLAVA, Negative heads frequently concentrate near the final layers (e.g., Layers 29–31), suggesting late-stage attention drift or feature overfitting.
    \item Certain heads (e.g., \texttt{L11 H28}, \texttt{L23 H30}) appear recurrently across multiple semantic tasks in InstructBLIP, potentially functioning as universal positive heads.
\end{itemize}

These results further substantiate our claim that LVLMs develop semantically differentiated reasoning pathways, and offer concrete targets for causal editing interventions such as Attention rescaling.

\paragraph{Behavioral Divergence of Positive and Negative Heads.}

To better quantify the behavioral differences between positive and negative heads, we compute the average spatial alignment of their attention distributions with ground-truth object bounding boxes. Specifically, for each head, we calculate the attention overlap ratio—defined as the sum of attention weights inside the object bounding box divided by the total attention mass. Results are averaged across the top-10 positive and negative heads per model and per dataset. As shown in Table~\ref{tab:attn_overlap}, positive heads show strong alignment (e.g., 83.8\% in LLaVA-Animal), while negative heads often misalign (as low as 7.1\%), typically attending to irrelevant background regions such as grass or walls. These findings quantitatively support the causal distinction between the two groups.

\begin{table}[t]
\centering
\resizebox{\linewidth}{!}{ 
\begin{tabular}{lcc}
\toprule
\textbf{Subset} & \textbf{Number of Samples} & \textbf{Example Question} \\
\midrule
Popular      & 3,000  & Is there a chair in the image? \\
Adversarial  & 3,000  & Is there a man in the image? \\
Random       & 2,910  & Is there a bottle in the image? \\
\bottomrule
\end{tabular}
}
\vspace{0.5em}
\caption{Overview of POPE subsets. Each subset contains binary object presence questions designed for different evaluation purposes.}
\label{tab:pope-subsets}
\end{table}

\begin{table}[t]
\centering
\resizebox{\linewidth}{!}{ %
\begin{tabular}{lcc}
\toprule
\textbf{Question Type} & \textbf{Number of QA Pairs} & \textbf{Percentage} \\
\midrule
Object     & 69,753  & 59.3\% \\
Number     & 16,594  & 14.1\% \\
Color      & 16,201  & 13.8\% \\
Location   & 15,136  & 12.8\% \\
\midrule
\textbf{Total} & \textbf{117,684} & \textbf{100\%} \\
\bottomrule
\end{tabular}
}
\vspace{0.5em}
\caption{Distribution of question types in COCO-QA. The dataset is dominated by object-related questions.}
\label{tab:cocoqa-question-types}
\end{table}

\section {Dataset and Task Details for Causal Attention Head Evaluation}
\label{sec:dataset-Causal Attention Head Evaluation}

\subsection { POPE: A Benchmark for Object Hallucination Evaluation}

The POPE (Polling-based Object Probing Evaluation) dataset ~\cite{li2023evaluating} is a diagnostic benchmark specifically designed to evaluate large vision-language models (LVLMs) on object hallucination in recognition tasks. Built on images from the MS-COCO dataset, POPE focuses on probing model robustness and bias under distribution shifts via object-level binary questions.

The POPE dataset consists of three subsets, each targeting a specific evaluation goal:

\begin{itemize}
  \item \textbf{Popular}: Contains frequently occurring objects in pretraining corpora. This subset is used to test the model's reliance on common object priors.

  \item \textbf{Adversarial}: Introduces misleading distractors or co-occurrence biases to evaluate model robustness in challenging scenarios.

  \item \textbf{Random}: Includes randomly sampled object-question pairs, serving as a control group for unbiased evaluation.
\end{itemize}

Each image is paired with multiple binary questions, typically in the format: “Is there a \texttt{<object>} in the image?”, where \texttt{<object>} is a specific noun.

In our experiments, we directly transfer attention heads identified from the GQA dataset to POPE, and compare their performance with those re-identified from 10\% and 20\% of POPE’s training samples.

The statistics and examples of each subset are shown in Table~\ref{tab:pope-subsets}.

\subsection {COCO-QA: A Dataset for Open-Ended Visual Question Answering}

The COCO-QA (COCO Question Answering) dataset~\cite{lu2016hierarchical} is an automatically generated visual question answering (VQA) dataset based on images from the MS-COCO Captions dataset. It is designed to evaluate image understanding and language grounding in an open-ended QA setting.

\vspace{0.5em}
\noindent\textbf{Dataset Structure.}  
COCO-QA consists of:

\begin{itemize}
  \item \textbf{Images:} Derived from the MS-COCO dataset, with approximately 123,287 unique images.
  \item \textbf{Question-Answer Pairs:} A total of 117,684 QA pairs, each linked to a COCO image. Questions are automatically generated from image captions using syntactic and semantic transformations. Answers are typically single words or short phrases.
\end{itemize}

\begin{table*}[!t]
\centering
\small
\renewcommand{\arraystretch}{1.2}
\resizebox{0.95\textwidth}{!}{
\begin{tabular}{llccccccc}
\toprule
\textbf{Model} & \textbf{Ablation} & \multicolumn{2}{c}{\textbf{Attribute}} & \multicolumn{3}{c}{\textbf{Object}} & \multicolumn{2}{c}{\textbf{Relation}} \\
\cmidrule(lr){3-4} \cmidrule(lr){5-7} \cmidrule(lr){8-9}
& & \textbf{Color} & \textbf{Material} & \textbf{Animal} & \textbf{Vehicle} & \textbf{Indoor} & \textbf{Action} & \textbf{Spatial} \\
\midrule
\multirow{5}{*}{LLaVA} 
& original & 76.13 & 78.84 & 96.45 & 90.29 & 83.80 & 88.21 & 77.85 \\
& w/o negative & 79.73 & 83.77 & 97.01 & 92.70 & 88.58 & 92.89 & 85.33 \\
& w/o positive & 74.40 & 75.62 & 95.33 & 89.20 & 79.64 & 86.44 & 73.18 \\
& random remove & 76.00 & 78.92 & 96.36 & 90.11 & 83.13 & 88.08 & 77.79 \\
& rescaling & \textbf{80.27} & \textbf{85.31} & \textbf{97.10} & \textbf{92.87} & \textbf{89.77} & \textbf{93.29} & \textbf{88.92} \\
\midrule
\multirow{5}{*}{InstructBLIP} 
& original & 77.53 & 81.38 & 96.17 & 90.06 & 87.14 & 90.88 & 87.64 \\
& w/o negative & 80.47 & 85.08 & 97.38 & 93.10 & 92.78 & 95.19 & 95.59 \\
& w/o positive & 75.47 & 80.23 & 94.86 & 87.30 & 80.45 & 81.17 & 71.90 \\
& random remove & 77.53 & 81.15 & 96.07 & 90.00 & 87.14 & 91.86 & 86.51 \\
& rescaling & \textbf{81.27} & \textbf{86.84} & \textbf{97.48} & \textbf{93.85} & \textbf{92.83} & \textbf{96.03} & \textbf{96.77} \\
\bottomrule
\end{tabular}}
\caption{Accuracy (\%) of LLaVA and InstructBLIP on VQA tasks categorized by task type: Attribute (Color, Material), Object (Animal, Vehicle, Indoor), and Relation (Action, Spatial). The table reports performance under different attention head editing strategies.}
\label{tab:head-edit-comparison_appendix}
\end{table*}

\vspace{0.5em}
\noindent\textbf{Question Types.}  
COCO-QA questions are categorized into four semantic types:

\begin{itemize}
  \item \textbf{Object:} Identifying objects in the image (e.g., ``What is on the table?'').
  \item \textbf{Number:} Counting entities (e.g., ``How many people are there?'').
  \item \textbf{Color:} Recognizing object attributes (e.g., ``What color is the car?'').
  \item \textbf{Location:} Understanding spatial relationships (e.g., ``Where is the cat?'').
\end{itemize}

As shown in Table~\ref{tab:cocoqa-question-types}, object-related questions constitute the majority of the dataset, followed by number, color, and location types.

\vspace{0.5em}
\noindent\textbf{Subset Selection.}  
In our experiments, we select the \textit{Object}, \textit{Color}, and \textit{Location} subsets for evaluation. These three categories exhibit clear semantic meanings and align closely with our defined semantic reasoning types: object recognition, attribute identification, and spatial relation understanding. We exclude the \textit{Number} subset due to its relatively ambiguous semantics and lack of direct correspondence to our reasoning framework.

\begin{table*}[ht]
\centering
\begin{tabular}{llcccc}
\toprule
Category   & Baseline        & Mean $\Delta$(pp)(\%) & SD(\%) & t     & p-value  \\ \midrule
Attribute  & original        & 7.92           & 2.96   & 84.79 & ***      \\
           & random remove   & 7.80           & 3.52   & 70.20 & ***      \\
           & w/o positive    & 10.55          & 3.95   & 84.61 & ***      \\
           & w/o negative    & 2.93           & 1.87   & 49.68 & ***      \\
Object     & original        & 3.01           & 1.51   & 63.16 & ***      \\
           & random remove   & 3.19           & 2.06   & 49.06 & ***      \\
           & w/o positive    & 4.78           & 2.27   & 66.75 & ***      \\
           & w/o negative    & 1.93           & 1.18   & 51.81 & ***      \\
Relation   & original        & 9.57           & 3.82   & 79.40 & ***      \\
           & random remove   & 10.43          & 3.93   & 84.08 & ***      \\
           & w/o positive    & 13.62          & 4.39   & 98.29 & ***      \\
           & w/o negative    & 3.65           & 2.14   & 54.00 & ***      \\
Overall    & original        & 6.62           & 3.61   & 100.49& ***      \\
           & random remove   & 6.75           & 3.73   & 99.17 & ***      \\
           & w/o positive    & 9.86           & 4.46   & 121.18& ***      \\
           & w/o negative    & 2.88           & 1.89   & 83.49 & ***      \\
\bottomrule
\end{tabular}
\caption{Performance difference $\Delta$(pp) between the attention head rescaling method and baselines. Statistical significance is indicated with asterisks. *** indicates p < 0.001 (extremely significant).}
\label{tab:head-edit-comparison-statistical-significance}
\end{table*}

\section{Additional Results on Semantic Attention Editing}
\label{sec:appendix-attention_editing-results}

To complement our main results on LLaVA, we report additional results of Attention head rescaling applied to the InstructBLIP model across both GQA-style semantic sub-tasks and the POPE dataset. These results demonstrate that head rescaling is also effective when applied to more instruction-tuned multimodal models, such as InstructBLIP.

\paragraph{Effectiveness of Attention Head Editing}
\label{sec:appendix-effectiveness}

Table~\ref{tab:head-edit-comparison_appendix} presents the results of systematically editing the identified key attention heads. We observe that for both LLaVA and InstructBLIP, attention head editing brings significant performance improvements across all semantic levels. These results confirming that our method effectively enhances the semantic understanding capabilities of vision-language models.

\paragraph{Statistical Significance Analysis}
\label{sec:appendix-statistical_significance_analysis}

To assess whether the performance gains are statistically significant, we applied bootstrap resampling \cite{koehn2004statistical} by generating 1,000 folds, each consisting of 100 test cases randomly sampled from the original test sets, with each set emphasizing a specific type of semantic change (i.e., object, relationship, or attribute). For each fold, we calculated the performance difference ($\Delta$(pp)) between our attention head rescaling method and each of the four baselines, defined as $\Delta$(pp) = rescaling - baseline.

For each semantic property, we obtained 1,000 performance difference values per method comparison and conducted paired t-tests \cite{student1908probable} to assess the statistical significance between each pair of methods. Table~\ref{tab:head-edit-comparison-statistical-significance} shows the results based on the LLaVA model. Note that the “Overall” results are calculated based on all test folds across the three semantic properties for each method comparison.

As shown in the Table~\ref{tab:head-edit-comparison-statistical-significance}, our rescaling strategy achieves statistically significant improvements over all four baselines across every category of the LLaVA model. Paired t-tests further show the robustness of these results, with all p-values being extremely small (p < 0.001), indicating a high level of statistical significance.

\begin{table}[htbp]
\centering
\renewcommand{\arraystretch}{1.25}
\resizebox{0.48\textwidth}{!}{
\begin{tabular}{lcccccc}
\toprule
\textbf{Category} & \textbf{Initial} & \textbf{10\%} & \textbf{20\%} & \textbf{30\%} & \textbf{50\%} & \textbf{Full} \\
\midrule
Attribute & 79.46 & 80.48 & 82.11 & 83.05 & 83.73 & \textbf{84.06} \\
Object    & 91.12 & 92.78 & 93.57 & 94.37 & 94.61 & \textbf{94.72} \\
Relation  & 89.26 & 91.42 & 93.56 & 95.27 & 96.10 & \textbf{96.40} \\
\midrule
Average   & 86.61 & 88.23 & 89.75 & 90.90 & 91.48 & \textbf{91.73} \\
\bottomrule
\end{tabular}
}
\caption{Accuracy (\%) of \textbf{InstructBLIP} under different sample proportions for rescaling, grouped by task categories.}
\label{tab:soft_reweighting_instructblip_overall}
\end{table}

\begin{table}[htbp]
\centering
\renewcommand{\arraystretch}{1.25}
\resizebox{\linewidth}{!}{
\begin{tabular}{lcccccc}
\toprule
\textbf{Dataset} & \textbf{Initial} & \textbf{10\%} & \textbf{20\%} & \textbf{30\%} & \textbf{50\%} & \textbf{Full} \\
\midrule
Color     & 77.53  & 78.80  & 79.60  & 80.33  & 81.07  & \textbf{81.27} \\
Material  & 81.38  & 82.15  & 84.62  & 85.77  & 86.38  & \textbf{86.84} \\
Indoor    & 87.14  & 89.48  & 90.63  & 92.02  & 92.50  & \textbf{92.83} \\
Animal    & 96.17  & 96.72  & 97.20  & \textbf{97.48}  & \textbf{97.48}  & \textbf{97.48} \\
Vehicle   & 90.06  & 92.13  & 92.87  & 93.62  & \textbf{93.85}  & \textbf{93.85} \\
Action    & 90.88  & 92.79  & 93.52  & 94.89  & 95.86  & \textbf{96.03} \\
Spatial   & 87.64  & 90.05  & 93.59  & 95.64  & 96.34  & \textbf{96.77} \\
\bottomrule
\end{tabular}
}
\vspace{0.5em}
\caption{Accuracy (\%) of \textbf{InstructBLIP} under different sample proportions for attention head rescaling across the seven semantic tasks defined in Section~\ref{sec:Semantic benchmark}.}
\label{tab:soft_reweighting_instructblip_detail}
\end{table}

\begin{table}[htbp]
\centering
\renewcommand{\arraystretch}{1.25}
\resizebox{\linewidth}{!}{
\begin{tabular}{lcccccc}
\toprule
\textbf{Dataset} & \textbf{Initial} & \textbf{10\%} & \textbf{20\%} & \textbf{30\%} & \textbf{50\%} & \textbf{Full} \\
\midrule
Color     & 76.13  & 77.73  & 79.80  & 79.93  & 80.00  & \textbf{80.27} \\
Material  & 78.84  & 81.23  & 82.54  & 84.46  & 85.00  & \textbf{85.31} \\
Indoor    & 83.80  & 87.48  & 89.67  & 89.77  & \textbf{89.82}  & 89.77 \\
Animal    & 96.45  & 97.01  & \textbf{97.10}  & \textbf{97.10}  & \textbf{97.10}  & \textbf{97.10} \\
Vehicle   & 90.29  & 92.76  & 92.82  & 92.84  & \textbf{92.87}  & \textbf{92.87} \\
Action    & 88.21  & 90.58  & \textbf{93.29}  & 93.22  & 93.26  & \textbf{93.29} \\
Spatial   & 77.85  & 83.44  & 88.62  & 88.82  & 88.87  & \textbf{88.92} \\
\bottomrule
\end{tabular}
}
\vspace{0.5em} 
\caption{Accuracy (\%) of \textbf{LLaVA} under different sample proportions for rescaling across various VQA tasks.}
\label{tab:soft_reweighting_detail_llava}
\end{table}

\paragraph{Data Dependence Analysis}
\label{sec:appendix-data-dependence}

To assess the data dependence of our method, we randomly sample 10\%--50\% of the experimental data for key head identification and editing, then test on the full set. Tables~\ref{tab:soft_reweighting_instructblip_overall}, ~\ref{tab:soft_reweighting_instructblip_detail}, and~\ref{tab:soft_reweighting_detail_llava} show the performance of InstructBLIP and LLaVA under different data proportions.

For InstructBLIP, even with only 10\% of the training data, our method achieves performance close to full-data supervision across all task categories. Table~\ref{tab:soft_reweighting_instructblip_detail} further provides detailed performance of InstructBLIP on seven specific semantic tasks, demonstrating the data efficiency and general applicability of our method.

For LLaVA, Table~\ref{tab:soft_reweighting_detail_llava} shows that even in low-resource scenarios (using only 10\% of the data), our method can achieve comparable performance, highlighting the advantages of attention head rescaling in practical applications.

\paragraph{Generalizability Evaluation}
\label{sec:appendix-generalizability}

We evaluate the generalizability of our editing method on two out-of-distribution (OOD) benchmarks: POPE and COCOQA.

\begin{table}[t]
\centering
\small
\renewcommand{\arraystretch}{1.15}
\resizebox{0.48\textwidth}{!}{
\begin{tabular}{llccc}
\toprule
\textbf{Task} & \textbf{Metric} & \textbf{Initial} & \textbf{Edit\_POPE} & \textbf{Edit\_GQA} \\
\midrule
\multirow{3}{*}{Popular} 
& Acc     & 0.842 & \textbf{0.861} & 0.857 \\
& Recall  & 0.731 & 0.782 & \textbf{0.795} \\
& F1      & 0.823 & \textbf{0.843} & 0.841 \\
\midrule
\multirow{3}{*}{Adversarial} 
& Acc     & 0.830 & 0.846 & \textbf{0.848} \\
& Recall  & 0.731 & \textbf{0.787} & 0.783 \\
& F1      & 0.811 & 0.828 & \textbf{0.829} \\
\midrule
\multirow{3}{*}{Random} 
& Acc     & 0.852 & \textbf{0.872} & 0.869 \\
& Recall  & 0.731 & 0.786 & \textbf{0.790} \\
& F1      & 0.835 & \textbf{0.853} & 0.854 \\
\bottomrule
\end{tabular}
}
\caption{Performance of \textbf{InstructBLIP} on POPE under different attention head rescaling strategies. "Initial" refers to the performance without any intervention. "Edit\_POPE" refers to applying head rescaling based on POPE. "Edit\_GQA" denotes using heads and scores discovered on GQA.}
\label{tab:instructblip-pope}
\end{table}

\begin{table}[ht]
\centering
\small
\begin{tabular}{lcc}
\toprule
\textbf{Model} & \textbf{Initial} & \textbf{Edit\_GQA} \\
\midrule
\multicolumn{3}{l}{\textit{Color Task}} \\
InstructBLIP & 65.54 & \textbf{68.49} \\
LLaVA        & 80.48 & \textbf{82.35} \\
\midrule
\multicolumn{3}{l}{\textit{Object Task}} \\
InstructBLIP & 78.40 & \textbf{80.12} \\ 
LLaVA        & 85.15 & \textbf{86.40} \\ 
\midrule
\multicolumn{3}{l}{\textit{Location Task}} \\
InstructBLIP & 67.39 & \textbf{69.01} \\
LLaVA        & 61.84 & \textbf{62.98} \\
\bottomrule
\end{tabular}
\caption{Performance of InstructBLIP and LLaVA models on the \textit{Color}, \textit{Object}, and \textit{Location} sub-tasks of COCOQA before and after applying attention head rescaling. Accuracy is reported in percentage.}
\label{tab:cocoqa-color-object-location}
\end{table}

\paragraph{Performance on POPE Dataset}
Table~\ref{tab:instructblip-pope} demonstrates the performance of InstructBLIP on the POPE dataset. We compare three strategies: (1) no attention editing ("Initial"), (2) head rescaling using POPE data ("Edit\_POPE"), and (3) using attention heads and importance scores identified from GQA ("Edit\_GQA"). The results show that even without re-analysis on the target dataset, attention head rescaling significantly improves VLM's performance, demonstrating good cross-task generalization capability of our method.

\paragraph{Performance on COCOQA Dataset}
Table~\ref{tab:cocoqa-color-object-location} shows the direct comparison between GQA-edited VLMs and unedited baselines on the COCOQA dataset, which contains open-ended questions covering color, object, and location categories. The results show consistent performance improvements across all tasks for the edited models, further proving the generalizability of our method.

In summary, these appendix results comprehensively demonstrate the effectiveness and generalizability of the Attention head rescaling method across different models, data quantities, and tasks, providing strong support for the arguments presented in the main body of this paper.

\section{Computational Resources}

All experiments were conducted on a single machine equipped with two NVIDIA Tesla V100 GPUs (32GB memory), an Intel(R) Xeon(R) Gold 6230R CPU running at 2.10GHz (8 cores), and 62GB of RAM.

Our interpretability analyses, including activation patching, attention head attribution, and visual perturbation generation, were implemented using PyTorch and Hugging Face Transformers. 

\begin{table*}[ht]
\centering
\small
\label{tab:semantic-perturbation-examples}
\begin{tabular}{|>{\centering\arraybackslash}m{4.5cm}
                |>{\centering\arraybackslash}m{2.5cm}
                |>{\centering\arraybackslash}m{2.2cm}
                |>{\centering\arraybackslash}m{2.5cm}
                |>{\centering\arraybackslash}m{2.2cm}|}
\toprule
\textbf{Question} & \textbf{Original Image} & \textbf{Original Prediction} & \textbf{Perturbed Image} & \textbf{Perturbed Prediction} \\
\midrule

Is the shirt blue? & 
\includegraphics[width=2.2cm]{Appendix/color.jpg} & 
\texttt{No} & 
\includegraphics[width=2.2cm]{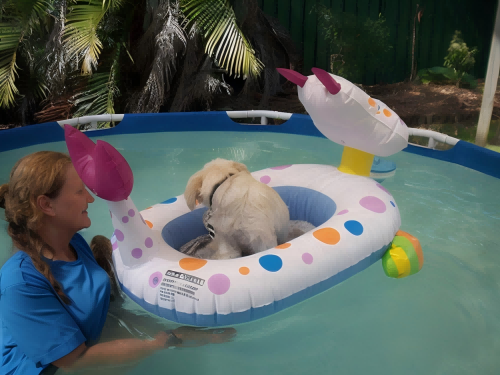} & 
\texttt{Yes} \\

\midrule

Is the chair made of wood? & 
\includegraphics[width=2.2cm]{Appendix/material.jpg} & 
\texttt{Yes} & 
\includegraphics[width=2.2cm]{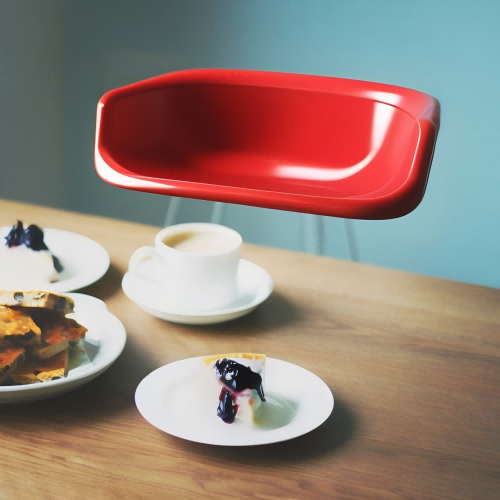} & 
\texttt{No} \\

\midrule

Is there a mirror in this picture? & 
\includegraphics[width=2.2cm]{Appendix/indoor.jpg} & 
\texttt{Yes} & 
\includegraphics[width=2.2cm]{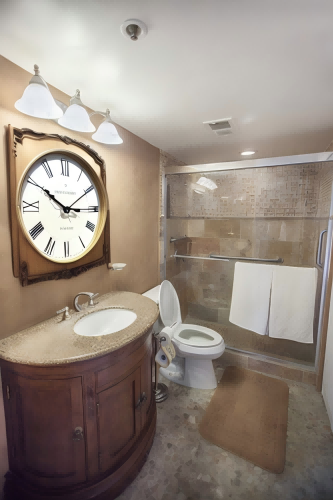} & 
\texttt{No} \\

\midrule

Is there a dog in this picture? & 
\includegraphics[width=2.2cm]{Appendix/animal.jpg} & 
\texttt{No} & 
\includegraphics[width=2.2cm]{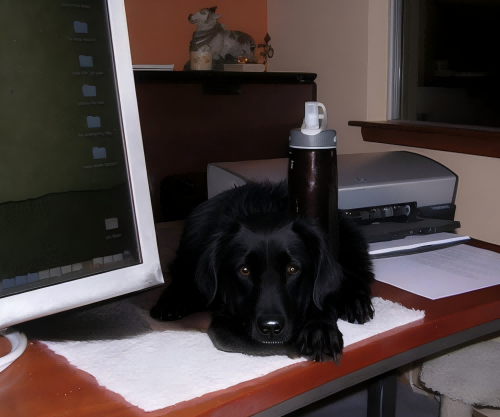} & 
\texttt{Yes} \\

\midrule

Is there a bike in this picture? & 
\includegraphics[width=2.2cm]{Appendix/vehicle.jpg} & 
\texttt{Yes} & 
\includegraphics[width=2.2cm]{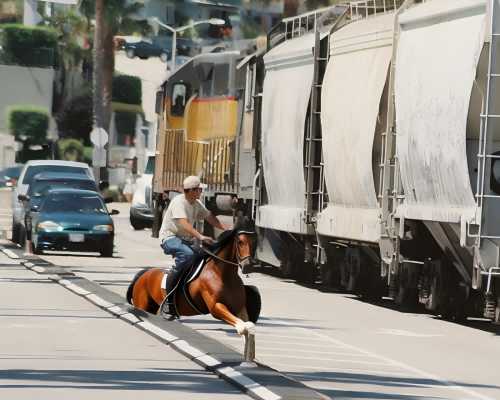} & 
\texttt{No} \\

\midrule

Is the man right of the horse? & 
\includegraphics[width=2.2cm]{Appendix/spatial.jpg} & 
\texttt{No} & 
\includegraphics[width=2.2cm]{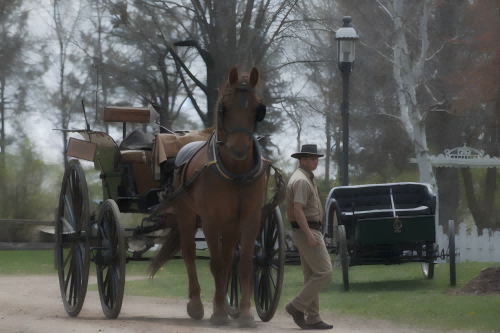} & 
\texttt{Yes} \\

\midrule

Is the woman holding the skis? & 
\includegraphics[width=2.2cm]{Appendix/action.jpg} & 
\texttt{Yes} & 
\includegraphics[width=2.2cm]{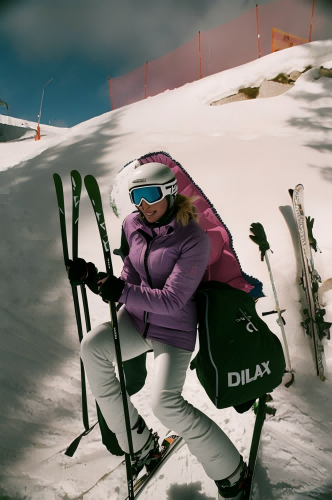} & 
\texttt{No} \\

\bottomrule
\end{tabular}
\caption{Qualitative examples of semantic perturbation on seven VQA tasks using LLaVA. Each row presents the VQA question, original image and model prediction, and the perturbed image with the updated prediction.}
\end{table*}

\end{document}